%% file: paper.tex
\documentclass[]{bytedance_seed}

\usepackage[toc,page,header]{appendix}

\usepackage{minitoc}
\usepackage{cleveref} 
\input{macro}
\usepackage{titletoc}
\usepackage[ruled,vlined]{algorithm2e}
\usepackage[symbol]{footmisc}

\usepackage{multirow}
\usepackage{booktabs}
\usepackage{tabularx}
\usepackage{colortbl}
\usepackage[most]{tcolorbox}
\usepackage{setspace}
\usepackage{graphicx}
\usepackage{fancyvrb}
\usepackage{svg}
\usepackage{array}
\usepackage{listings}
\usepackage{pifont}
\usepackage{float}
\usepackage{amsmath}
\usepackage{amssymb}
\usepackage{amsfonts}
\usepackage{bm}
\usepackage{wrapfig}
\usepackage{mathtools}

\usepackage[T1]{fontenc}
\usepackage{adjustbox}   
\usepackage{hyperref}
\usepackage[edges]{forest}
\usepackage{subcaption}
\usepackage{xcolor}
\usepackage{enumitem}
\usepackage{courier}

\definecolor{new_green}{HTML}{56C596}
\definecolor{new_sharrow_blue}{HTML}{4DA5D9}
\definecolor{new_blue}{HTML}{3C88C6}
\definecolor{new_purple}{HTML}{6A0DAD}  
\usepackage{pifont}
\usepackage{bbding}
\usepackage{fontawesome}
\definecolor{my_green}{RGB}{40,154,121}
\definecolor{my_yellow}{RGB}{255,165,0}
\definecolor{my_red}{RGB}{176,46,46}

\newcommand{\method}[0]{\textsc{Mole-Syn}\xspace}

\tcbset{
  takeaway/.style={
    width=\linewidth,
    top=8pt,
    bottom=4pt,
    colback=blue!6!white,
    colframe=black,
    colbacktitle=black,
    enhanced,
    center,
    attach boxed title to top left={yshift=-0.1in,xshift=0.15in},
    boxed title style={boxrule=0pt,colframe=white,},
  }
}

\newtcolorbox{TakeawayBox}[2][]{takeaway,title=#2,#1}

\tcbset{
  prompt/.style={
    width=\linewidth,
    top=8pt,
    bottom=4pt,
    colback=yellow!6!white,
    colframe=black,
    colbacktitle=black,
    enhanced,
    center,
    attach boxed title to top left={yshift=-0.1in,xshift=0.15in},
    boxed title style={boxrule=0pt,colframe=white,},
  }
}

\newtcolorbox{PromptBox}[2][]{prompt,title=#2,#1}

\title{The Molecular Structure of Thought: Mapping the Topology of Long Chain-of-Thought Reasoning}

\author[1,2]{Qiguang Chen}
\author[1,\dagger]{Yantao Du}
\author[1]{Ziniu Li}
\author[2]{Jinhao Liu}
\author[3]{Songyao Duan}
\author[3]{\\Jiarui Guo}
\author[4]{Minghao Liu}
\author[5]{Jiaheng Liu}
\author[3]{Tong Yang}
\author[6]{Ge Zhang}
\author[7,\dagger]{\\Libo Qin}
\author[2,\dagger]{Wanxiang Che}
\author[1]{Wenhao Huang}

\affiliation[1]{ByteDance Seed China}
\affiliation[2]{LARG, SCIR, Harbin Institute of Technology}
\affiliation[3]{Peking University}
\affiliation[4]{2077AI Foundation}
\affiliation[5]{Nanjing University}
\affiliation[6]{M-A-P}
\affiliation[7]{Central South University}

\abstract{
\input{sections/000abstract}
}

\date{Jan 15, 2026}
\correspondence{\url{{qgchen,car}@ir.hit.edu.cn}, \url{duyantao@bytedance.com},\url{lbqin@csu.edu.cn}}
\begin{document}

\maketitle
\begin{figure*}[!h]
    \centering
    \vspace{-20pt}
    \includegraphics[width=0.92\textwidth]{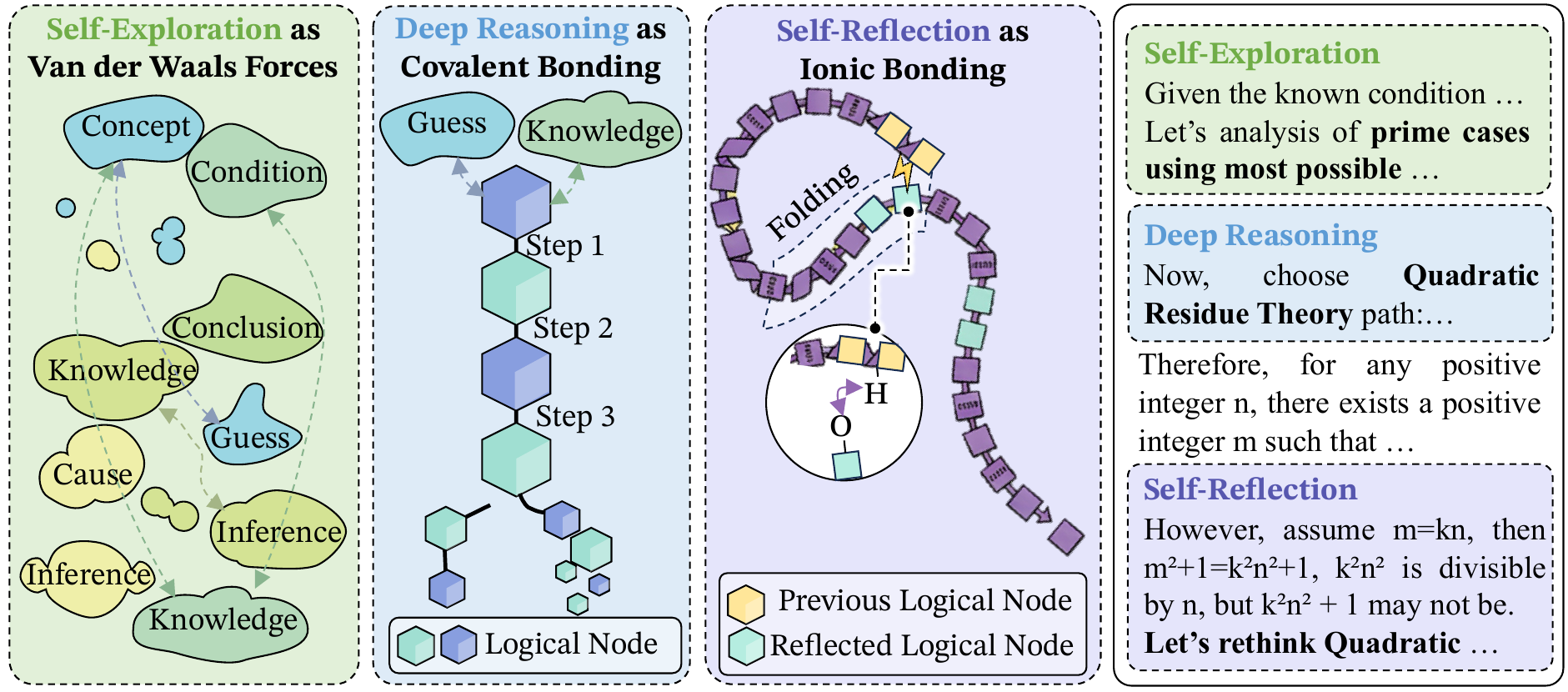}
    \caption{The hypothesis that stable molecular structure in Long CoT arises from three key “chemical” bonds.}
    \label{fig:main}
\end{figure*}
\input{sections/010intro}

\input{sections/020preliminary}
\input{sections/030hypothesis}

\input{sections/040verification}
\input{sections/050feature}
\input{sections/060synthetic}

\input{sections/070function}

\input{sections/080protect}
\input{sections/090related}

\input{sections/100conclusion}

\bibliographystyle{unsrtnat}
\bibliography{main}

\clearpage

\beginappendix

\input{sections/appendix}

\end{document}

%% file: macro.tex
\usepackage{natbib}
\usepackage{latexsym}

\usepackage{url}
\usepackage{amssymb}
\usepackage[utf8]{inputenc}
\usepackage{microtype}
\usepackage{booktabs}
\usepackage{pifont} 
\usepackage{multirow}
\usepackage{makecell}
\usepackage{paralist}
\usepackage{xspace}
\usepackage{color}
\usepackage{xcolor}
\usepackage{colortbl}
\usepackage{adjustbox}
\usepackage{hyperref} 
\usepackage[edges]{forest}
\usepackage{tikz} 
\usepackage{caption}
\usepackage{amsfonts}

\hypersetup{
    colorlinks,
    linkcolor={blue!80!black},
    citecolor={blue!80!black},
}
\tikzset{
    root/.style =             {align=center, text width=1cm, rounded corners=3pt, line width=0.3mm, fill=gray!10, draw=gray!80, font=\small},
    demographic/.style =         {align=center, text width=1.8cm, rounded corners=3pt, line width=0.3mm, fill=blue!10, draw=blue!80, font=\footnotesize},
    demographic_work/.style =    {align=center, text width=10cm, rounded corners=3pt, line width=0.3mm, fill=blue!10, draw=blue!0, font=\footnotesize},
    character/.style =         {align=center, text width=1.8cm, rounded corners=3pt, line width=0.3mm, fill=red!10, draw=red!80, font=\footnotesize},
    character_work/.style =    {align=center, text width=10cm, rounded corners=3pt, line width=0.3mm, fill=red!10, draw=red!0, font=\footnotesize},
    personalization/.style =           {align=center, text width=1.8cm, rounded corners=3pt, line width=0.3mm, fill=cyan!10, draw=cyan!80, font=\footnotesize},
    personalization_work/.style =      {align=center, text width=10cm, rounded corners=3pt, line width=0.3mm, fill=cyan!10, draw=cyan!0, font=\footnotesize},
    risk/.style =         {align=center, text width=1.8cm, rounded corners=3pt, line width=0.3mm, fill=orange!10, draw=orange!80, font=\footnotesize},
    risk_work/.style =    {align=center, text width=10cm, rounded corners=3pt, line width=0.3mm, fill=orange!10, draw=orange!0, font=\footnotesize},
}

\usepackage{CJK}

%% file: sections/000abstract.tex
\begin{abstract}{}
    Large language models (LLMs) often fail to learn effective long chain-of-thought (Long CoT) reasoning from human or non-Long-CoT LLMs imitation. To understand this, we propose that effective and learnable Long CoT trajectories feature stable molecular-like structures in unified view, which are formed by three interaction types: Deep-Reasoning (covalent-like), Self-Reflection (hydrogen-bond-like), and Self-Exploration (van der Waals-like). Analysis of distilled trajectories reveals these structures emerge from Long CoT fine-tuning, not keyword imitation. We introduce Effective Semantic Isomers and show that only bonds promoting fast entropy convergence support stable Long CoT learning, while structural competition impairs training. Drawing on these findings, we present \method, a distribution-transfer-graph method that guides synthesis of effective Long CoT structures, boosting performance and RL stability across benchmarks.
\end{abstract}

%% file: sections/010intro.tex
\section{Introduction}

Recently, large language models (LLMs) have excelled on diverse reasoning tasks via explicit chain-of-thought (CoT) rationales~\citep{wei2022chain,wang2022self,kojima2022large,qin2024large}. Yet, they struggle to cold-start from instruction-tuned or base models into Long CoT models requiring extended multi-step reasoning~\citep{chen2025towards,chen-etal-2025-unveiling-key}. Notably, \citet{du2025the} shows that humans generate Long CoT rationales without imitating DeepSeek-R1~\citep{guo2025deepseek}. Our preliminary studies reveal that standard supervised fine-tuning and distillation from human or Instruction LLM rationales (using randomly sampled Long CoT examples) fail to reliably instill these skills in LLMs. Models often lose coherence over long trajectories or fail to transfer patterns to novel tasks. This prompts a key question:

\begin{quotation}
{\begin{center}
    \textbf{How do Large Language Models learn and represent effective Long Chain-of-Thought?}
\end{center}}
\end{quotation}

\begin{figure*}[t]
    \centering
    \includegraphics[width=0.98\textwidth]{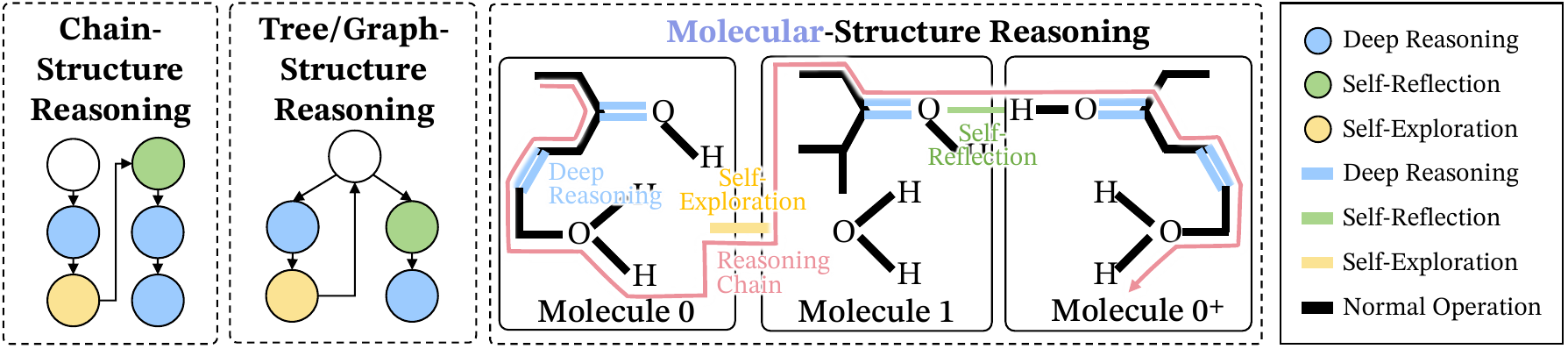}
    \caption{Comparison of prior chain- or tree-like structures and our molecular structure. Reasoning starts from Mole. 0, uses \textbf{deep reasoning} on strongly related structures, then employs \textbf{self-exploration} for new logic in Mole. 1. When meet errors, reasoning utilize \textbf{self-reflection} to guide chain to optimized Mole. 0$^{+}$.}
    
    \label{fig:intro}
\end{figure*}

To explain this, we posit that they acquire the organization of reasoning trajectories. As shown in Figure~\ref{fig:intro}, prior studies model these as \emph{logic nodes} in sequences or trees of steps. Yet, our analysis of Long CoT across strong reasoning models reveals a stable distribution of three core behaviors across tasks and architectures: Deep-Reasoning, Self-Reflection, and Self-Exploration~\citep{chen2025towards}, which node-centric views fail to capture.

This finding triggers a molecular-inspired, distributional view: we model behavior-labeled \emph{logic edges} as interaction bonds and examine how their global \textbf{molecular-like structure} ensures long-horizon reasoning stability. Specifically, Deep-Reasoning forms dense local clusters of coupled deductions, like covalent bonds; Self-Reflection creates long-range corrective links to prior steps, like hydrogen bonds; and Self-Exploration forges weak bridges between distant clusters, like van der Waals forces. Thus, high-quality Long CoT arises from the {\textbf{stable composition and arrangement of these bond types}}, guiding effective learning\footnote{Note: C, H, and O atom references are analogies for molecular structure only.}.

In this framework, we define semantic isomers as Long CoT trajectories that solve the same tasks and visit similar semantic regions but differ in behavior distributions and transitions. We demonstrate that multiple near-optimal semantic isomers exist per task family, but mixing stable isomers from different strong teachers destabilizes learning, degrading performance and behavior distributions despite matched token statistics. This structurally explains why combining heterogeneous Long CoT traces often fails, beyond token-level distillation.

Building on this perspective, we propose \method, a structure-aware synthesis framework that first estimates a behavior transition graph from strong reasoning models and then transfers only this behavioral structure to cheaper instruction LLMs via controlled trajectory synthesis, instead of directly copying teacher outputs. This decouples structural transfer from model-specific surface form, enables the generation of Long CoT data that match target behavior distributions from scratch, and yields consistent gains in both Long CoT performance and RL stability across six  benchmarks.

After that, we analyze the shaping function of each bond in the Long CoT structure. Deep Reasoning bonds encode core logical flow, Self-Reflection bonds support folding pathways to previous steps, and Self-Exploration bonds reinforce long-range consistency checks, enabling targeted bond distributions. Moreover, we discuss why a deteriorated molecular structure is hard to restore, which helps explain how private LLMs protect Long CoT structures from distillation-based imitation. Methods such as summarization and reasoning compression can disrupt Long CoT structure, limiting unauthorized replication of internal reasoning processes.

In summary, our contributions are as follows:
\begin{itemize}[leftmargin=16pt, itemsep=0pt, topsep=0pt]
\item We model Long CoT as a molecular structure with 3 bonds: deep-reasoning (covalent), self-reflection (hydrogen-bond), and self-exploration (van der Waals), to understand its effective learning.
\item We identify effective Semantic Isomers for Long CoT learning, where only entropy-convergent bonds enable stable learning, while competing structures destabilize learning.
\item We introduce \method, which uses distribution-transfer graphs to synthesize these structures, improving Long CoT and stabilizing RL across 6 benchmarks.
\end{itemize}

%% file: sections/020preliminary.tex
\section{Preliminary: Cold-start LLMs for Long Chain-of-Thought}
First, we validate whether LLMs exhibit valid Long CoT trajectories suitable for Long CoT learning. We investigate three data sources: distillation from strong reasoning LLMs, distillation from weak instruction LLMs with in-context learning (ICL), and fine-tuning on human reasoning traces.\vspace{-5pt}

\begin{figure*}[t]
    \centering
    \includegraphics[width=0.96\textwidth]{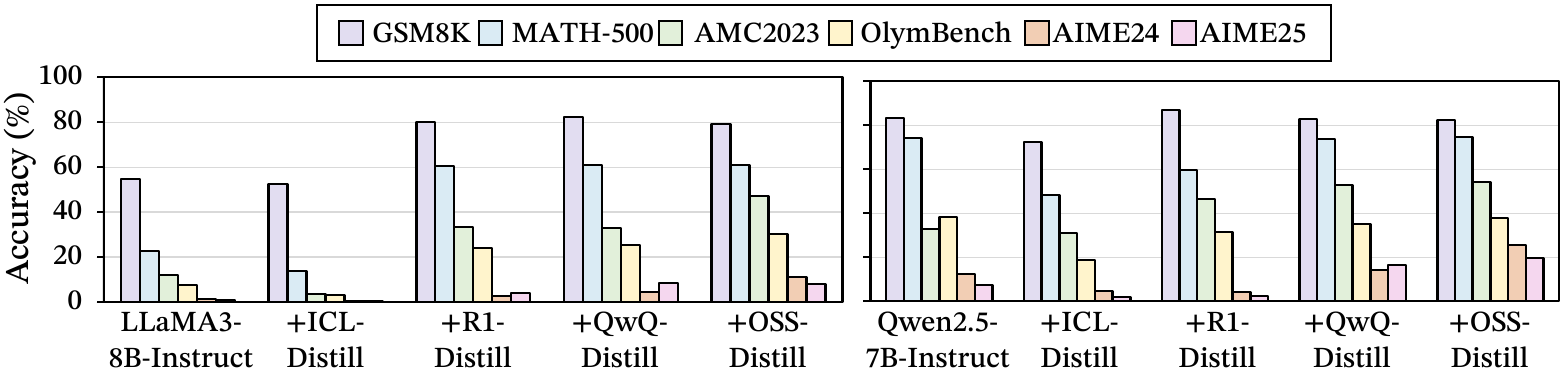}
    
    \caption{The failure of distillation from weak instruction LLMs with ICL and Human-annotated reasoning traces to acquire Long CoT structures, compared to successful distillation from strong reasoning LLMs. See Appendix Figure~\ref{fig:comparison-full} for the full result.}
    
    \label{fig:comparison}
\end{figure*}
\begin{wrapfigure}{r}{0.47\textwidth}
    \centering
    \includegraphics[width=0.46\textwidth]{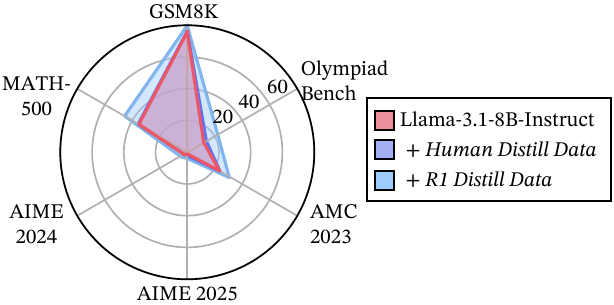}
    \caption{Performance comparison between human-annotated reasoning traces (\textit{+ Human Distill Data}) and R1 distilled reasoning traces (\textit{+ R1 Distill Data}).}
    \label{fig:human}
    \vspace{-5pt}
\end{wrapfigure}

\paragraph{\textbf{Only distillation from strong reasoning LLMs works.}}
To identify effective data sources, we curated a synthetic set of reasoning traces from three sources. As shown in Figure~\ref{fig:comparison}, only distillation from strong reasoning LLMs enables target models to learn and retain Long CoT structure, improving performance on benchmarks requiring extended reasoning. These results indicate that only high-quality traces reliably support both learning and use of Long CoT structures.\vspace{-8pt}

\paragraph{\textbf{Distillation from randomly selected ICL by weak instruct LLMs does not work.}}
Instruction LLMs without explicit reasoning training lag behind reasoning-specialized models. We test whether LLMs can acquire Long CoT structure by distilling from an instruction LLM using randomly selected ICL demonstrations that emulate Long CoT reasoning. As shown in Figure~\ref{fig:comparison}, performance drops sharply. Instruction LLMs can only mimic short CoT traces ($\thicksim$6–8 steps) and fail to extend exploration while preserving intermediate steps and trace coherence. This degradation indicates a limitation of ICL-based distillation rather than robust long-chain imitation.\vspace{-8pt}

\paragraph{\textbf{Even human-annotated Long-CoT-like traces fail.}}
Inspired by \citet{du2025the}, we test whether human step-by-step solutions can induce long CoT. We collect human solutions for complex reasoning tasks and fine-tune LLMs on them. Figure~\ref{fig:human} shows that human-trace training does not reproduce the long-CoT gains from distilling strong reasoning models, suggesting that human solutions aid local problem solving but may not reliably encode abstractions for long-horizon reasoning distributions.

\begin{TakeawayBox}{Takeaway 1}
Distillation from strong reasoning LLMs effectively imparts long CoT structures, while ICL from weak instruction models and fine-tuning on human traces yield limited gains. This underscores the importance of high-quality reasoning exemplars for robust long-chain learning in LLMs.
\end{TakeawayBox}

%% file: sections/030hypothesis.tex
\section{Hypothesis: Stable ``Molecular Structure'' in Long CoT}

To understand these phenomena, as shown in Fig.~\ref{fig:main} \& \ref{fig:intro}, effective Long CoT connects nodes in logical space through reasoning behaviors, forming a stable macromolecular structure with mutually supportive components from a global perspective.

We formalize a Long CoT trace as a behavior-directed graph \(G = (V, E)\), where each node \(v \in V\) represents a reasoning step or edge \(s = (u, v) \in E\) is annotated with a behavior type \(s\rightarrow b \in \{\mathcal{D}, \mathcal{R}, \mathcal{E}\}\). For a trace corpus \(\mathcal{C}\), we estimate the behavior transition \(P_{\mathcal{C}}(b' \mid b)\) over consecutive edges and the marginal distribution \(\pi_{\mathcal{C}}(b)\).
Empirically, strong reasoning teachers produce stable \((P_{\mathcal{C}}, \pi_{\mathcal{C}})\) across models and tasks.
Specifically, these include the following three major bonds\footnote{Math definitions and proofs are in Appendix~\ref{append:math}.}:\vspace{-8pt}

\paragraph{\textbf{Deep Reasoning as Covalent Bonds}}
Deep reasoning forms the bone of the thought process, analogous to covalent bonds defining a molecule's primary chain. It encodes strong logical dependencies (Step A must justify Step B), maintaining direction and continuity; breaking this bone undermines the following steps and destabilizes the answer. By contrast, ``Normal Operation'' corresponds to stable local bonds within each step, capturing routine computation and direct semantic expression.\vspace{-8pt}

\paragraph{\textbf{Self-Reflection as Hydrogen Bonds}}
Reflection is a key stabilizer. As proteins gain stability when chains fold and form intra-chain hydrogen bonds, reasoning stabilizes when later steps (e.g., Step 100) test, revise, or reinforce earlier premises (e.g., Step 10). These long-range links constrain drift and hallucination, turning a long sequence into a more self-consistent structure. If later checks fail to align with earlier commitments, the reasoning cannot ``fold,'' indicating a structural logical error.

\paragraph{\textbf{Self-Exploration as Van der Waals Forces}}
Exploration resembles transient Van der Waals interactions: it supports abductive and inductive moves by enabling low-commitment associations in semantic space, where concepts can drift, combine, and be probed before stronger constraints are enforced.

%% file: sections/040verification.tex
\section{Verification: Molecular Structure}

\begin{figure*}[t]
    \centering
    \includegraphics[width=0.98\textwidth]{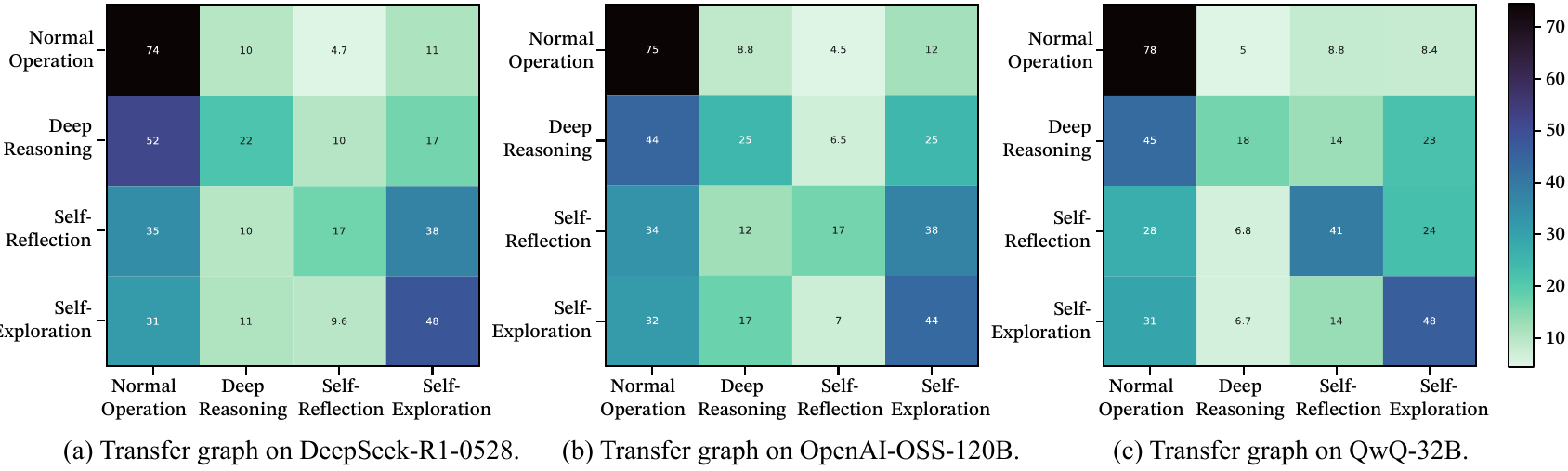}
    \caption{Transfer graph on three different models. Pearson correlation coefficients across models are all greater than 0.9 (p<0.001), when sampling examples > 2,000, transfer graphs will become stable and get over 0.95 Pearson correlation between different sampling sizes.}
    
    \label{fig:transfer}
\end{figure*}

\subsection{Stable Bond Distribution in Long CoT}
To address this, we first verify whether effective Long CoT traces show stable, macromolecular-like organization across models and tasks. As shown in Fig.~\ref{fig:transfer}, traces from multiple LLMs across diverse tasks yield Pearson correlations exceeding 0.9 (p<0.001) for over 2,000 samples. These transfer graphs stabilize with correlations above 0.95 across sampling sizes. This indicates that effective Long CoT structures rely on robust motifs: different models recover similar reasoning topologies across tasks, whereas simple human simulation or ICL cannot emulate the global bond distribution.

\subsection{SFT actually learns these bond structures rather than keywords.}
For the SFT analysis, we consider Llama-3.1-8B-Base that is pre-trained but not instruction-tuned on Long CoT data;
Then, we consider Llama-3.1-8B-Base trained on R1-distilled data as Think-SFT model obtained by supervised fine-tuning enriched with Long CoT traces.\vspace{-8pt}

\begin{figure*}[t]
    \centering
    \includegraphics[width=0.98\textwidth]{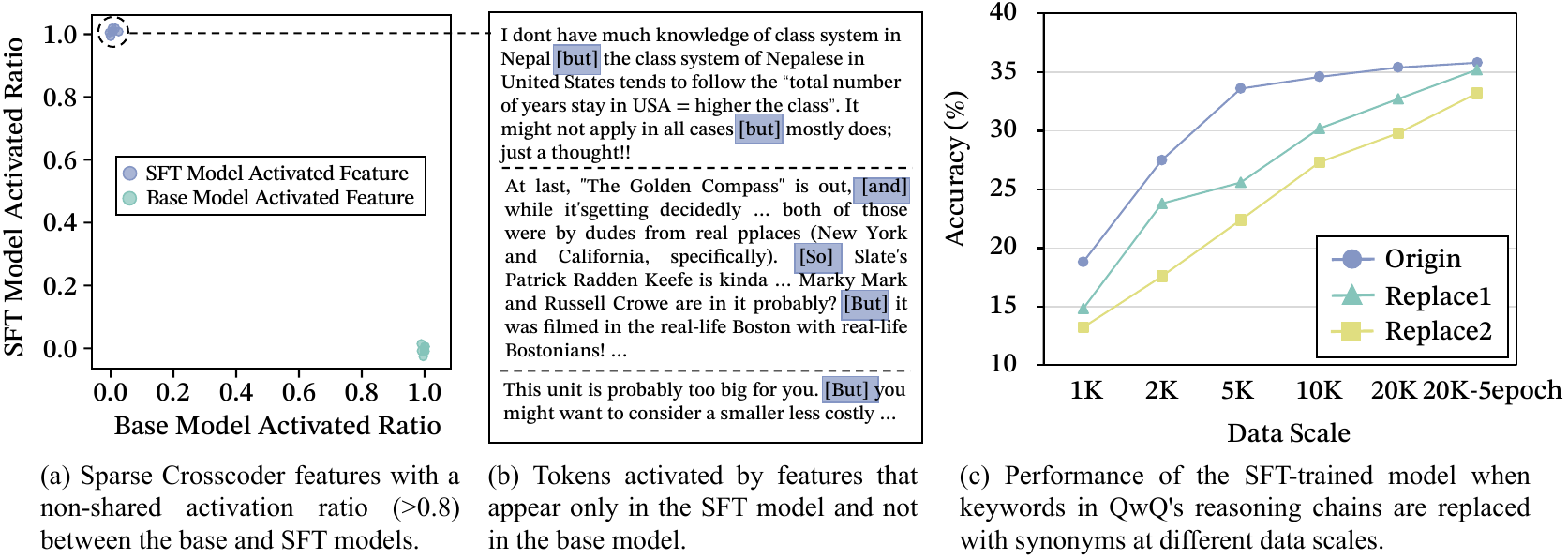}
    \caption{The learned features in Long CoT supervised fine-tuning.}
    \label{fig:learning}
\end{figure*}

From a representation-geometry perspective, the sparse auto-encoder analysis shows that Long CoT behavior in the SFT model is concentrated in a small set of discourse-control structures rather than being uniformly distributed across tokens. As shown in Figure~\ref{fig:learning} (a, b), we train a cross-coder sparse auto-encoder that jointly models hidden states from the base model and the Think-SFT model, and then identify features whose activation on think tokens is at least threefold higher than their average activation rate. Within this highly activated subset, many features are predominantly driven by a few connective keywords, such as  \textit{``Maybe''}, \textit{``But / so''}, and \textit{``Alternatively''}, indicating that the SFT process has carved out dedicated latents for managing local hypothesis revision, contrastive moves, and branch selection in long CoT traces.

Based on previous analyses, we argue that models learn the characteristic reasoning behaviors these keywords represent, not the keywords themselves. Following \citet{chen2025towards}, we define three categories of Long CoT reasoning behaviors and test this claim using two training datasets derived from QwQ distillation data. In the first, we replace each keyword with one of four alternative variants. In the second, we remove all keywords while preserving the reasoning trajectories. We then fine-tune identical LLMs on each dataset and evaluate their Long CoT reasoning performance.

As shown in Figure~\ref{fig:learning} (c), explicit keywords like "wait" accelerate learning but are not essential. Models trained without keywords, or with arbitrary alternatives, achieve comparable reasoning performance given sufficient training, provided the underlying reasoning behaviors remain intact. This reveals a fundamental insight: LLMs internalize reasoning structure of reasoning rather than surface lexical cues. Consequently, training data should prioritize the distribution of reasoning behaviors over specific keyword choices to effectively enhance model reasoning capabilities.

However, a key open question remains: do these bonds drive Long CoT structure learning, and if so, \textit{\textbf{why do explicit human imitation or random ICL distillation of these markers often fail?}}

\begin{figure*}[t]
    \centering
    \includegraphics[width=0.96\textwidth]{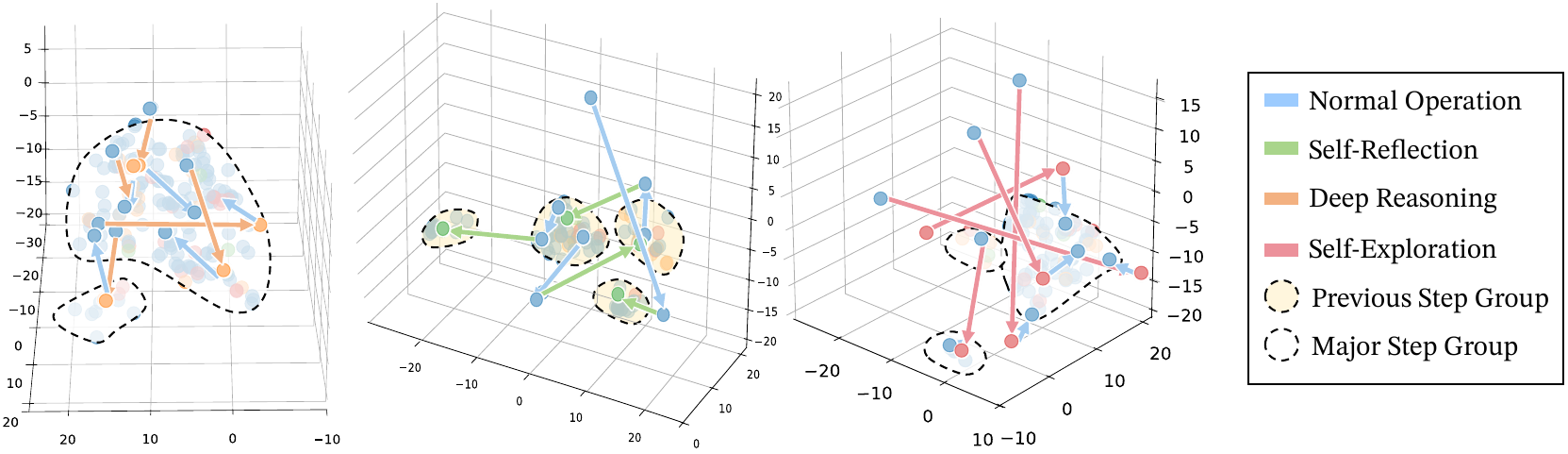}
    \caption{Verification of the ``logical folding'' structure on embeddings from Qwen2.5-32B-Instruct and t-SNE-based low-dimensional representations from the OpenAI-OSS-120B-generated reasoning process.}
    
    \label{fig:validation}
\end{figure*}

\begin{figure*}[t]
    \centering
    \includegraphics[width=0.76\textwidth]{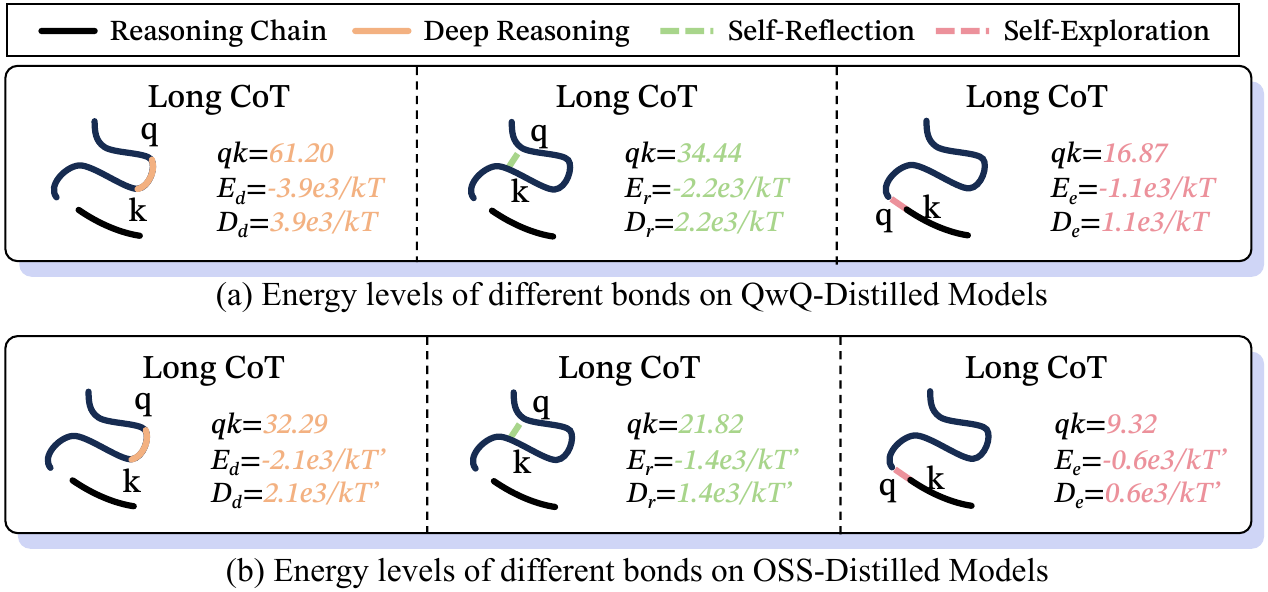}
    
    \caption{Energy levels of different bonds across two distilled data.}
    
    \label{fig:attention}
\end{figure*}

\subsection{``Logical Bonding-Folding'' Structure}

To test the hypothesis that Long CoT is analogous to macro-molecular folding, as shown in Fig.~\ref{fig:validation}, we analyzed the topology of CoT in a 3D semantic space. Each trajectory edge was classified as reflection, deep reasoning, or exploration, and its geometric properties were quantified.\vspace{-8pt}

\paragraph{\textbf{Deep-Reasoning stabilizes logical cluster by Covalent Bonding.}}
Modeled as covalent bonds, as shown in Fig.~\ref{fig:validation} (a), deep reasoning bonds mainly increase local connectivity, forming stable subdomains. After deep reasoning, 72.56\% of steps remained within a group distance of less than 3 in the semantic space (generally, group-group distance $>$5.6).\vspace{-8pt}

\paragraph{\textbf{Self-Reflection drives strong folding to previous steps by Hydrogen Bonding.}}
Analogous to hydrogen bonds, as shown in Fig.~\ref{fig:validation} (b), self-reflection transitions fold later steps back onto earlier, semantically similar clusters rather than extending the chain linearly. Quantitatively, 81.72\% of reflection steps reconnected to a previously formed cluster with high semantic similarity.\vspace{-8pt}

\paragraph{\textbf{Self-Exploration gently links different long-distance clusters by Van der Waals Forces.}}
In contrast to local stabilization (Fig.~\ref{fig:validation} (c)), exploration transitions act as loose links between otherwise separated clusters. They show much larger step-to-step distances, with an average trajectory length of 5.32 in the 3D t-SNE projection.
Together, these results suggest that effective Long CoT reasoning is not a simple linear chain; instead, it forms a folded, domain-structured topology consistent with the ``logical folding'' hypothesis.

\subsection{Attention $\Leftrightarrow $ Energy Level of Bonds}
In physical chemistry, the behavior probability with energy level $E_i$ at temperature \(T\) follows the Boltzmann distribution:
\begin{equation}
    P(\text{state}_i) = \frac{\exp\!\big(-E_i / k_B T\big)}{\sum_j {\exp\!\big(-E_j / k_B T\big)}}.
    \label{eq:energy}
\end{equation}
In Transformers, the attention weight \(\alpha_{ij}\) of the \(i\)-th token to the \(j\)-th token is:
\begin{equation}
\alpha_{ij} = \frac{\exp\!\big(q_i \cdot k_j / \sqrt{d_k}\big)}{\sum_l \exp\!\big(q_i \cdot k_l / \sqrt{d_k}\big)}.
\end{equation}
The correspondence follows by defining attention energy \((E)\) \(\leftrightarrow\) \(( - q \cdot k )\), implies lower $E_{ij}$ and thus higher behavior probability.

Formally, our analysis only relies on the observation that attention weights form a Gibbs–Boltzmann distribution over negative logits. We therefore use the term “energy” to denote reparameterized logits and study how their expectations differ across behavior types.

\begin{table*}[t]
\centering
\resizebox{0.96\textwidth}{!}{%
\begin{tabular}{lcccccccc}
\toprule
    \textbf{Model} & \textbf{GSM8K} & \textbf{MATH-500} & \textbf{AIME2024} & \textbf{AIME2025} & \textbf{AMC2023} & \textbf{OlympiadBench} & \textbf{AVG} \\
    \midrule
    LLaMA-3.1-8B-Base~\citep{dubey2024llama} & 7.58 & 3.20 & 0.00 & 0.00 & 4.22 & 1.19 & 2.70 \\
    \texttt{\ \ + 20K R1-Distill-Data} & 63.38 & 30.60 & 0.21 & 0.42 & 14.22 & 8.30 & 19.52 \\
    \texttt{\ \ + 20K OSS-Distill-Data} & 75.89 & 54.20 & 4.38 & 6.46 & 37.34 & 23.85 & 33.69 \\
    \texttt{\ \ + 20K QwQ-Distill-Data} & 64.53 & 32.20 & 2.92 & 0.42 & 16.72 & 8.89 & 20.95 \\
    \midrule
    Llama-3.1-8B-Instruct~\citep{dubey2024llama} & 75.89 & 35.20 & 4.17 & 1.04 & 23.59 & 12.00 & 25.32 \\
    \texttt{\ \ + 20K R1-Distill-Data}& 79.91 & 60.60 & 2.50 & 3.88 & 33.13 & 23.85 & 33.98 \\
    \texttt{\ \ + 20K OSS-Distill-Data} & 79.00 & 60.80 & 10.83 & 7.71 & 47.03 & 30.22 & 39.27 \\
    \texttt{\ \ + 20K QwQ-Distill-Data} & 82.41 & 60.80 & 4.38 & 8.33 & 32.97 & 25.48 & 35.73 \\
\bottomrule
\end{tabular}
}

\caption{Results across six benchmarks. Full results are reported in Table~\ref{tab:full_distill_result} in the Appendix.}

\label{tab:distill_result}
\end{table*}
We then compare attention energy across Long CoT transition types. Fig.~\ref{fig:attention} shows distinct distributions for deep reasoning, reflection, and exploration. Deep reasoning exhibits the largest effective chemical-bond energy $D_d$. Reflection is intermediate, whereas exploration shows the weakest effective bond energy. This ordering and the relative proportions are consistent across models, supporting the hypothesis that a bond-like mechanism broadly links these reasoning behaviors.

\begin{TakeawayBox}{Takeaway 2}
\begin{itemize}[leftmargin=16pt, itemsep=0pt, topsep=0pt]
    \item Long CoT reasoning exhibits stable structures across models, with reasoning topologies converging.
    \item Semantic isomers, reasoning chains with identical concepts but different logical bonds, succeed or fail based on bond structure, not surface keywords.
    \item Three distinct logical bonds drive CoT structure: reflection folds back to prior clusters, deep reasoning creates stable local domains, and exploration bridges distant concepts, each with characteristic attention energy profiles matching Boltzmann-like distributions.
    \item SFT learns reasoning structure rather than surface keywords, determining Long CoT capability.
\end{itemize}
\end{TakeawayBox}

%% file: sections/050feature.tex
\section{Feature: Effective Semantic Isomers}

We now shift to examine the failure of surficial distillation: similar conceptual atoms can be linked by different bonds, producing completely different reasoning chains. We call these as \textbf{Semantic Isomers}. Mathematically, a semantic isomer of \(\mathcal{D}\) is any \(\mathcal{D}'\) whose \((P_{\mathcal{D}'}, \pi_{\mathcal{D}'})\) lies near \((P_{\mathcal{D}}, \pi_{\mathcal{D}})\) under a suitable divergence \(D(\cdot \,\|\, \cdot)\). Then, we will analyze how they form, are learned, and sometimes destabilize in Long CoT.

\subsection{Bond structure of Semantic Isomers is the key to Long-CoT learning}

\paragraph{\textbf{Well-structured Semantic Isomers can be effective for Long-CoT learning.}}
To assess whether well-structured Semantic Isomers can enhance Long-CoT learning, we constructed a Long CoT dataset by distilling from advanced reasoning LLMs. As shown in Table~\ref{tab:distill_result}, models exhibit consistent performance gains when trained on these variants (distribution correlation $\thicksim$0.9). This suggests that, within a certain range, multiple effective “allotropic”
variants of reasoning keys can exist.\vspace{-8pt}

\paragraph{\textbf{The model has multiple effective Semantic Isomers, and slight differences can significantly affect the results.}}
As shown in Table~\ref{tab:distill_result}, the structural correlation between R1- and OSS-derived chains can reach 0.95, yet for some models performance with R1-based chains drops by more than 10\%.
This gap indicates the presence of multiple near-optimal Semantic Isomers that are nonetheless fragile, where small changes in the distribution can cause large performance losses.\vspace{-8pt}

\begin{wrapfigure}{r}{0.32\textwidth}
    \centering
    \includegraphics[width=0.30\textwidth]{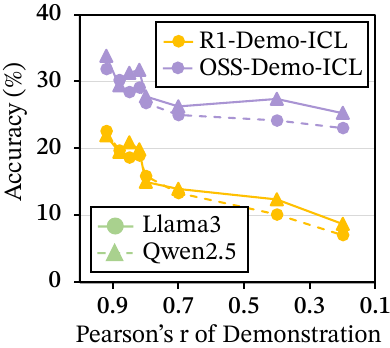}
    \caption{The performance of LLama-3.1-8B-Instruction under three different distillation setups from  ICL-augmented Qwen2.5-32B.}
    \vspace{-20pt}
    \label{fig:icl}
\end{wrapfigure}
\paragraph{\textbf{Simulating effective Semantic Isomer structures is the key to ICL distillation.}}
We analyze three ICL setups to simulate Long CoT chains by Qwen2.5-32B-Instruct: demonstrations chosen at random; demonstrations with closely aligned reasoning-key distributions (correlation \(\thicksim\)0.9); and demonstrations with clearly mismatched distributions (correlation \(< 0.8\)). As shown in Fig.~\ref{fig:icl}, substantial performance gains emerge only when demonstrations are constructed to match a specific target distribution of reasoning keys, thereby yielding an appropriate “allotrope” of that distribution.

\begin{figure*}[t]
    \centering
    \includegraphics[width=\textwidth]{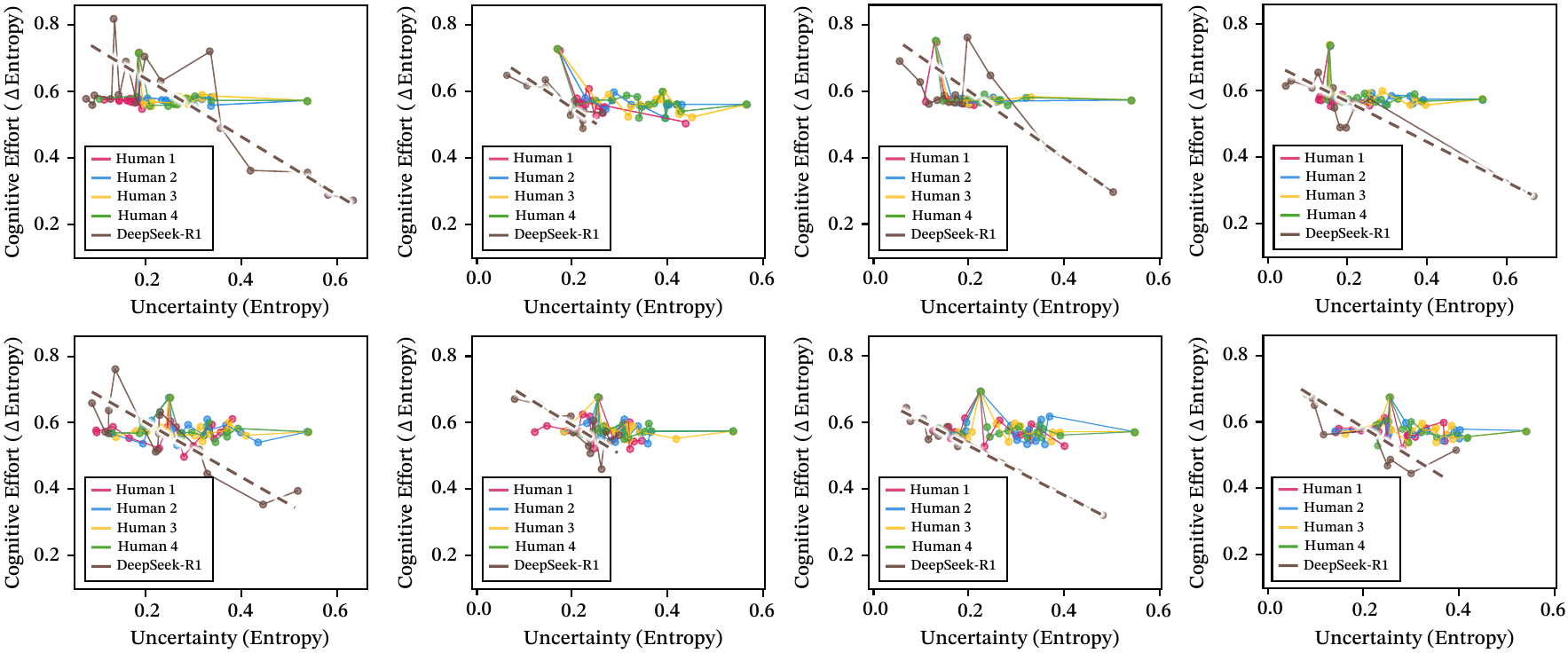}
    \caption{Information flow analysis for humans and reasoning models.}
    \label{fig:effect}
\end{figure*}

\begin{figure*}[t]
    \centering
    \includegraphics[width=0.98\textwidth]{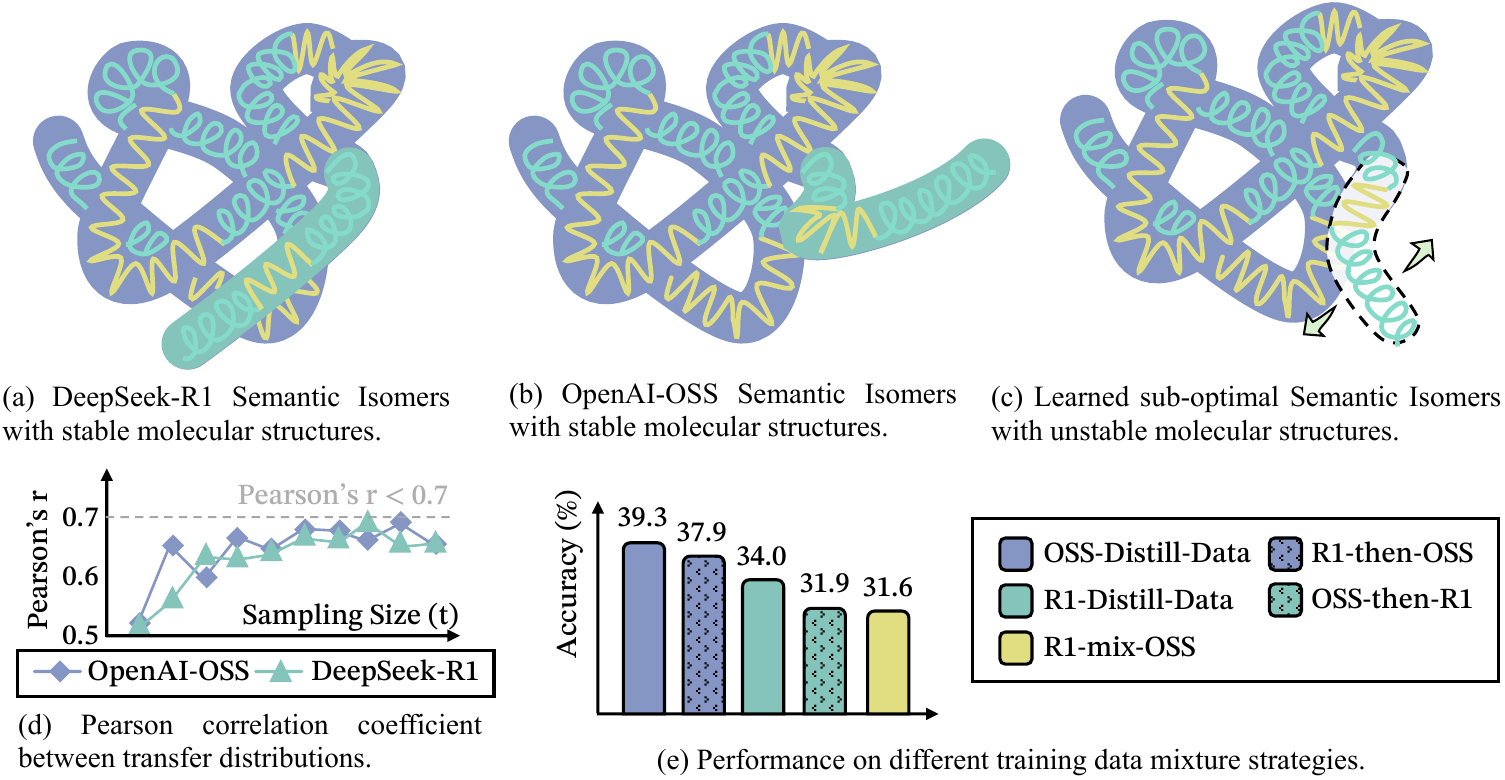}
    \caption{Conflict Learning between Two Stable Molecular Structures.}
    
    \label{fig:conflict}
\end{figure*}
\subsubsection{Not all bonds in isomers are effective.}

To clarify the nature of semantic isomers, we examine which bond structures yield effective reasoning configurations. We hypothesize that functional viability depends on specific bond distributions: despite sharing identical conceptual nodes, incompatible configurations disrupt information exchange. For instance, excessive exploration bonds cause fragmented reasoning, whereas overemphasized deep reasoning bonds create rigid chains unable to adapt to new inputs.Details are provided in Appendix~\ref{append:information}.\vspace{-8pt}

\paragraph{\textbf{Effective reasoning bond distribution influences the information divergence speed in reasoning dynamics.}}
To assess this, we compared the reasoning dynamics of R1 models with human cognition in an information phase space~\cite{chen2025universal}. Mechanistically, LLMs update by maximizing rewards and reducing entropy, whereas human reasoning is additionally constrained by semantic coherence and social feedback. Consequently, machine reasoning converges through accumulated gradient updates, whereas human reasoning stabilizes through iterative self-monitoring and social calibration.
As shown in Figure~\ref{fig:effect}, we tracked reasoning unfolding over extended chains in logical deduction tasks. Humans typically exhibit nearly uniform forward information gains (81.3\% of cases show changes $<0.1$), corresponding to a near-zero slope in phase space. In contrast, R1 models display accelerating informativeness (76.1\% of cases show absolute changes $>0.1$), progressing from low entropy to rapid convergence. These patterns indicate a fundamental difference in how R1 models and humans integrate information over time.\vspace{-8pt}

\paragraph{\textbf{Effective reasoning bonds cause metacognitive oscillation and alignment.}}
We identify the core rationale for this difference as a "metacognitive oscillation" in LLMs. Their responses alternate between high-entropy divergent exploration (slope \(> 0.6\), \(\Delta\)entropy \(> 0.05\)) and stable convergent validation, a dynamic less pronounced in the uniform entropy profiles of humans. Case studies (Figure~\ref{fig:effect}) confirm that R1 models utilize self-reflective revision to adjust reasoning paths against uncertainty. We hypothesize that aligning training objectives with these behavioral structure distributions can narrow the gap between machine and human reasoning dynamics.

\subsection{Conflict between Two Stable Structures}

Understanding how distinct reasoning structures interact reveals fundamental limits of complex cognitive systems. As shown in Fig.~\ref{fig:conflict} (a–c), forcibly fusing stable molecular isomers disrupts their backbone; analogously, combining incompatible reasoning frameworks breaks global logical coherence.\vspace{-8pt}

\paragraph{\textbf{Learning two heterogeneous stable structures at the same time will lead to structural chaos in the model.}}
As shown in Fig.~\ref{fig:conflict} (d), we test this by jointly activating two highly correlated (\(r \approx 0.9\)) reasoning chains from DeepSeek-R1 and OpenAI-OSS. Despite their similarity, co-activation prevents the model from converging to a single stable behavioral mode: it produces molecular bond distributions that fluctuate across samples and deviate from those characteristic of either OSS or R1. Consistent with this instability, the self-correlation of the jointly activated model does not exceed 0.8.\vspace{-8pt}

\paragraph{\textbf{This structural chaos leads to a significant decline in the performance of the model.}}
As shown in Fig.~\ref{fig:conflict} (e), joint activation also causes a marked drop in performance relative to either chain alone. This seemingly paradoxical effect indicates that structural compatibility, rather than mere statistical correlation, governs whether reasoning systems can coexist. The interference pattern suggests that the underlying cognitive architecture is rigid: without careful alignment, attempts to merge such systems yield fragmented, low-utility outputs instead of enhanced capability.

\begin{TakeawayBox}{Takeaway 3}
\begin{itemize}[leftmargin=16pt, itemsep=0pt, topsep=0pt]
    \item Well-structured isomers perform effectively when their reasoning-key distributions align closely, but small structural shifts cause fragility and sharp performance losses.
    \item Co-activating incompatible reasoning structures induces structural chaos that breaks coherence and degrades performance, proving that statistical similarity does not guarantee compatibility.
\end{itemize} 
\end{TakeawayBox}

%% file: sections/060synthetic.tex
\begin{table*}[t]
\centering
\resizebox{0.94\textwidth}{!}{
\begin{tabular}{lcccccccc}
\toprule
    \textbf{Model} & \textbf{GSM8K} & \textbf{MATH-500} & \textbf{AIME2024} & \textbf{AIME2025} & \textbf{AMC2023} & \textbf{OlympiadBench} & \textbf{AVG} \\
    \midrule
    LLaMA-3.1-8B-Base~\citep{dubey2024llama} & 7.58 & 3.20 & 0.00 & 0.00 & 4.22 & 1.19 & 2.70 \\
    \rowcolor{new_green!15}
    \texttt{\ \ + 20K Qwen-Distill-Data} & 62.47 & 29.40 & 0.00 & 0.00 & 12.81 & 6.81 & 18.58 \\
    \rowcolor{new_blue!15}
    \texttt{\ \ + 20K OSS-Distill-Data} & 75.89 & 54.20 & 4.38 & 6.46 & 37.34 & 23.85 & 33.69 \\
    \rowcolor{new_blue!15}
    \texttt{\ \ + 20K QwQ-Distill-Data} & 64.53 & 32.20 & 2.92 & 0.42 & 16.72 & 8.89 & 20.95 \\
    \rowcolor{new_purple!15}
    \texttt{\ \ + 20K OSS-\method{}} & 67.85  & 35.20  & 1.83  & 0.83  & 20.53  & 11.11  & 22.89 \\
    \rowcolor{new_purple!15}
    \texttt{\ \ + 20K QwQ-\method{}} & 66.41  & 35.00  & 2.08  & 0.63  & 20.16  & 10.37  & 22.44 \\
    \midrule
    Llama-3.1-8B-Instruct~\citep{dubey2024llama} & 75.89 & 35.20 & 4.17 & 1.04 & 23.59 & 12.00 & 25.32 \\
    \rowcolor{new_green!15}
    \texttt{\ \ + 20K Qwen-Distill-Data} & 76.50 & 39.80 & 4.38 & 1.04 & 25.63 & 19.70 & 27.84 \\
    \rowcolor{new_blue!15}
    \texttt{\ \ + 20K OSS-Distill-Data} & 79.00 & 60.80 & 10.83 & 7.71 & 47.03 & 30.22 & 39.27 \\
    \rowcolor{new_blue!15}
    \texttt{\ \ + 20K QwQ-Distill-Data} & 82.41 & 60.80 & 4.38 & 8.33 & 32.97 & 25.48 & 35.73 \\
    \rowcolor{new_purple!15}
    \texttt{\ \ + 20K OSS-\method{}} & 83.24 & 51.80 & 4.79 & 1.04 & 32.50 & 21.04 & 32.40  \\
    \rowcolor{new_purple!15}
    \texttt{\ \ + 20K QwQ-\method{}} & 84.31 & 50.20 & 5.21 & 1.67 & 32.34 & 20.00 & 32.29  \\
\bottomrule
\end{tabular}
}
\caption{Performance comparison across six benchmarks. Here, ``\raisebox{1pt}{\colorbox{new_purple!15}{ \rule[-0.2ex]{0pt}{1.0ex} }}'': distill from instruction LLMs+\method, ``\raisebox{1pt}{\colorbox{new_blue!15}{ \rule[-0.2ex]{0pt}{1.0ex} }}'': distill from reasoning LLMs, ``\raisebox{1pt}{\colorbox{new_green!15}{ \rule[-0.2ex]{0pt}{1.0ex} }}'': distill from instruction LLMs. See Table~\ref{tab:full_method_result} in Appendix for full results.}
\label{tab:synthesis_results}
\end{table*}

\begin{figure*}[t]
    \centering
    \includegraphics[width=0.98\textwidth]{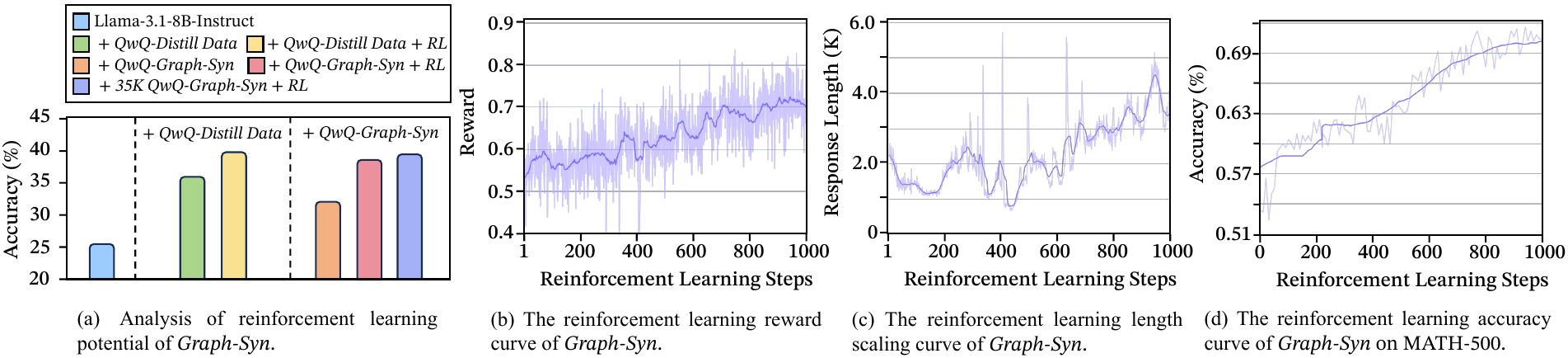}
    
    \caption{The continual improvement of RL performance with \method-initialized models. More details about the RL performance are in Table~\ref{tab:rl_results} in Appendix.}
    
    \label{fig:rl}
\end{figure*}

\section{Synthetic Chemistry: Synthesis Long CoT Molecules from Scratch}
LLMs may acquire advanced reasoning partly through exposure to explicit, structured Long CoT reasoning traces. However, it remains unclear how reliably such structures can be induced by prompting an instruction-tuned model, rather than obtained through distillation.\vspace{-8pt}

\paragraph{\textbf{\method Methodology.}}
To address this gap, we propose a synthetic framework, \method, that views targeted reasoning traces as macromolecular structures using only instruction LLMs. This method is a random walk on a transition probability graph in Figure~\ref{fig:transfer} composed of 4 reasoning behaviors from stronger reasoning LLMs that support Long CoT.\vspace{-8pt}

\paragraph{\textbf{\method can synthesize effective bond structures.}}
To test whether Long CoT capabilities can be learned from instruction LLMs, in Table~\ref{tab:synthesis_results}, we conduct training on \method generated data, which even achieves reasoning performance close to QwQ distillation. This suggests that instruction-driven synthesis can induce useful structural regularities, enabling lower-cost behavior transfer.\vspace{-8pt}

\begin{figure*}[t]
    \centering
    \includegraphics[width=0.98\textwidth]{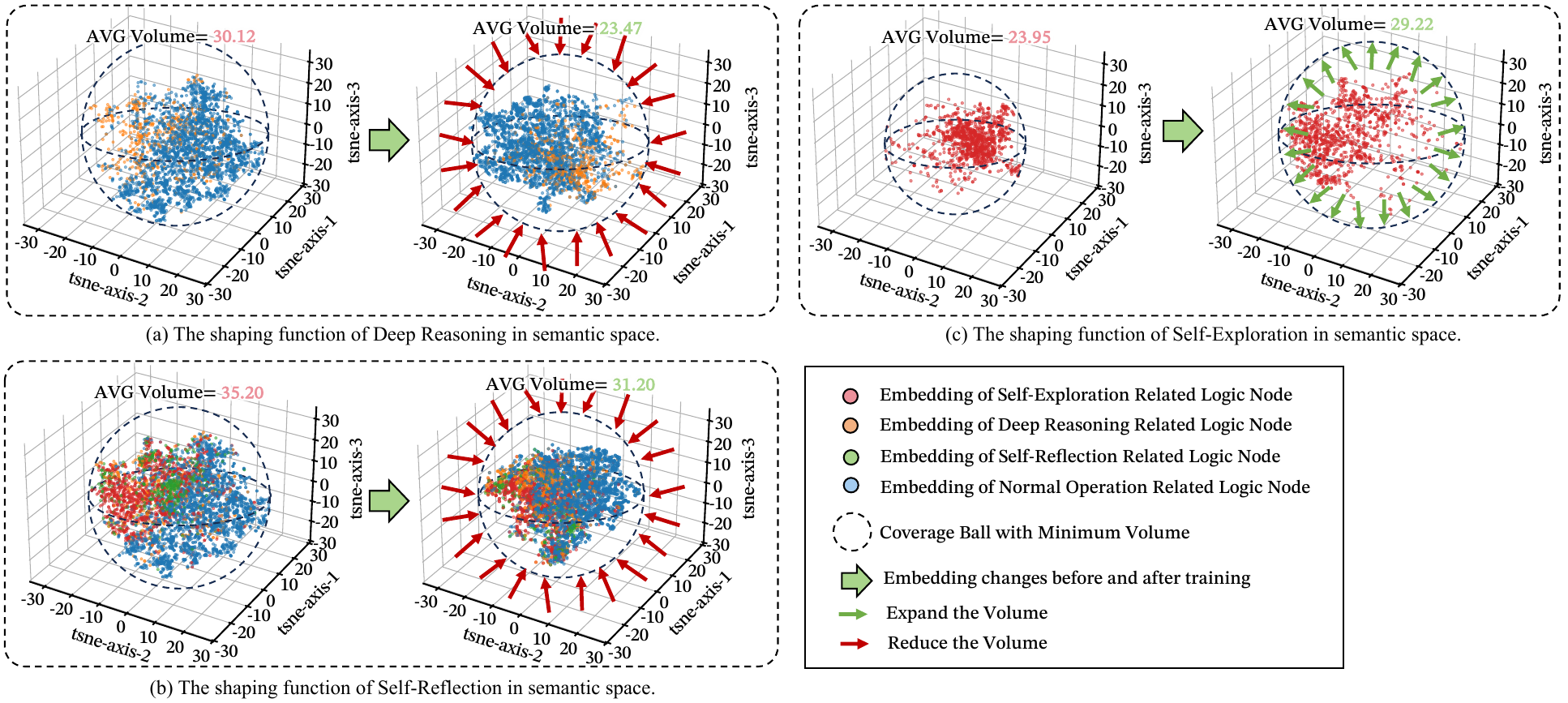}
    
    \caption{Roles of individual bonds in reasoning, inferred from semantic-space comparisons between Llama-3.1-8B-Instruct and Llama-3.1-8B-Instruct + QwQ-\method. Performance impact is shown in Fig.~\ref{fig:capability} (Appendix).}
    
    \label{fig:function}
\end{figure*}

\paragraph{\textbf{\method can further trigger stronger and continually improving RL.}}
We further evaluate \method-initialized LLMs' potential in reinforcement learning (RL). \method-initialized LLMs outperform those initialized from base LLMs. In Figure~\ref{fig:rl} (a), they show steadier fine-tuning gains, indicating stronger immediate reasoning and more reliable RL adaptation.
Moreover, Figures~\ref{fig:rl} (b-d) show that these gains persist over extended RL training, demonstrating durable benefits from synthesized Long CoT structures.
This effective use under RL supports their practical utility across diverse cognitive tasks.

\begin{TakeawayBox}{Takeaway 4}
\begin{itemize}[leftmargin=16pt, itemsep=0pt, topsep=0pt]
    \item \method{} successfully synthesizes Long CoT structures that match transition distributions from capable teacher models without requiring Long CoT distillation data.
    \item Transition-based Long CoT datasets achieve stable convergence and near-distillation performance, proving effective reasoning structures emerge purely from instruction-level synthesis at lower cost.
    \item Models initialized with synthesized Long CoT weights demonstrate superior and sustained RL performance gains, providing a robust foundation for continual learning in dynamic environments.
\end{itemize} 
\end{TakeawayBox}

%% file: sections/070function.tex
\section{Function: Shaping Function of Each Bond in Long CoT Structure}
Further, we analyze the shaping function of each bond in Long CoT structure. We suppose that LLM searches for an optimal semantic configuration parallels a protein’s descent along a folding funnel toward a low-energy native state (the solution).\vspace{-8pt}

\paragraph{\textbf{Deep Reasoning is densing major Structure Formation.}} As reasoning proceeds, Deep Reasoning drives Primary Structure formation by synthesizing a logical backbone, analogous to Covalent Bonding. As shown in Figure~\ref{fig:function} (a), Deep Reasoning densifies the core logical structure: the volume of the smallest covering ball in semantic space decreases by 22\% compared with baseline. This stage builds the answer’s skeleton, but it does not yet ensure global stability or correctness.\vspace{-8pt}

\paragraph{\textbf{Self-Reflection is densing and stabilize the global logics.}} After the bone is extended, Self-Reflection stabilizes the structure. Analogous to hydrogen bonds, reflection “folds” the logical chain by cross-linking distant nodes to test consistency, rather than adding new deductive steps. As shown in Figure~\ref{fig:function} (b), it consolidates the hydrophobic core and suppresses inconsistent branches, reducing the system volume from 35.2 to 31.2 and moving toward a stable, folded optimal state.\vspace{-8pt}

\paragraph{\textbf{Self-Exploration expand logical space.}} Self-Exploration broadens the set of feasible solutions and also increases variability and may introduce less consistent branches, so it improves coverage at the cost of reduced immediate stability. As shown in Figure~\ref{fig:function} (c), after learning Long CoT structure, exploration expands the exploration behavior in semantic space from 23.95 to 29.22.

\begin{TakeawayBox}{Takeaway 5}
Long CoT reasoning mimics protein folding through three stages:
Deep Reasoning densifies the logical backbone, Self-Exploration expands the search space to avoid local minima, and Self-Reflection converges toward a stable, optimized solution state in semantic space.
\end{TakeawayBox}

%% file: sections/080protect.tex
\begin{table*}[t]
\centering
\resizebox{\textwidth}{!}{
\begin{tabular}{lcccccccc}
\toprule
	\textbf{Model} & \textbf{GSM8K} & \textbf{MATH-500} & \textbf{AIME2024} & \textbf{AIME2025} & \textbf{AMC2023} & \textbf{OlympiadBench} & \textbf{AVG} \\
	\midrule
	Llama-3.1-8B-Instruct    & 75.89 & 35.20 & 4.17 & 1.04 & 23.59 & 12.00 & 25.32 \\
	\texttt{\ \ + 20K Gemini-Distill-Data}   & 54.89 & 22.60 & 1.04 & 0.63 & 12.03 & 7.41 & 16.43  \\
	\texttt{\ \ + 20K Claude4-Distill-Data}  & 63.76 & 35.80 & 1.04 & 0.83 & 23.44 & 13.48 & 23.06   \\
	\midrule
	Qwen-2.5-32B-Base        & 53.68 & 33.40 & 9.17 & 2.29 & 35.63 & 15.85 & 25.00  \\
	\texttt{\ \ + 20K Gemini-Distill-Data}   & 52.54 & 20.20 & 1.88 & 0.63 & 21.41 & 12.44 & 18.18  \\
	\texttt{\ \ + 20K Claude4-Distill-Data}  & 63.31 & 37.80 & 2.92 & 0.83 & 28.91 & 17.78 & 25.26  \\
	\midrule
	Qwen-2.5-32B-Instruct    & 93.71 & 81.00 & 15.60 & 14.17 & 69.84 & 42.22 & 52.76 \\
	\texttt{\ \ + 20K Gemini-Distill-Data}   & 63.68 & 32.80 & 15.00 & 2.92 & 35.63 & 19.11 & 28.19  \\
	\texttt{\ \ + 20K Claude4-Distill-Data}  & 76.88 & 54.80 & 17.71 & 13.96 & 55.00 & 30.96 & 41.55  \\
\bottomrule
\end{tabular}
}
\caption{Distillation Results from Gemini and Claude.}
\label{tab:distill}
\end{table*}
\begin{table*}[t]
\centering
\resizebox{\textwidth}{!}{
\begin{tabular}{lcccccccc}
\toprule
	\textbf{Model} & \textbf{GSM8K} & \textbf{MATH-500} & \textbf{AIME2024} & \textbf{AIME2025} & \textbf{AMC2023} & \textbf{OlympiadBench} & \textbf{AVG} \\
	\midrule
	Llama-3.1-8B-Instruct    & 75.89 & 35.20 & 4.17 & 1.04 & 23.59 & 12.00 & 25.32 \\
	\texttt{\ \ + 20K OSS-Summarized}   & 54.89 & 22.60 & 1.04 & 0.63 & 12.03 & 7.41 & 16.43  \\
	\texttt{\ \ + 20K QwQ-Summarized}  & 63.76 & 35.80 & 1.04 & 0.83 & 23.44 & 13.48 & 23.06\\
	\midrule
	Qwen-2.5-7B-Instruct & 83.24 & 74.00 & 12.50 & 7.08 & 22.66 & 38.07 & 39.59 \\
    \texttt{\ \ + 20K OSS-Summarized} & 82.34 & 72.60 & 12.50 & 6.46 & 21.41 & 27.70 & 37.17  \\
    \texttt{\ \ + 20K QwQ-Summarized} & 82.71 & 71.80 & 11.88 & 6.25 & 22.97 & 25.04 & 36.77  \\
	\bottomrule
\end{tabular}
}
\caption{Results from summarized Long CoT trajectories based on OpenAI-OSS and QwQ distilled trajectories.}
\label{tab:summarization}
\end{table*}

\begin{figure*}[t]
    \centering
    \includegraphics[width=0.98\textwidth]{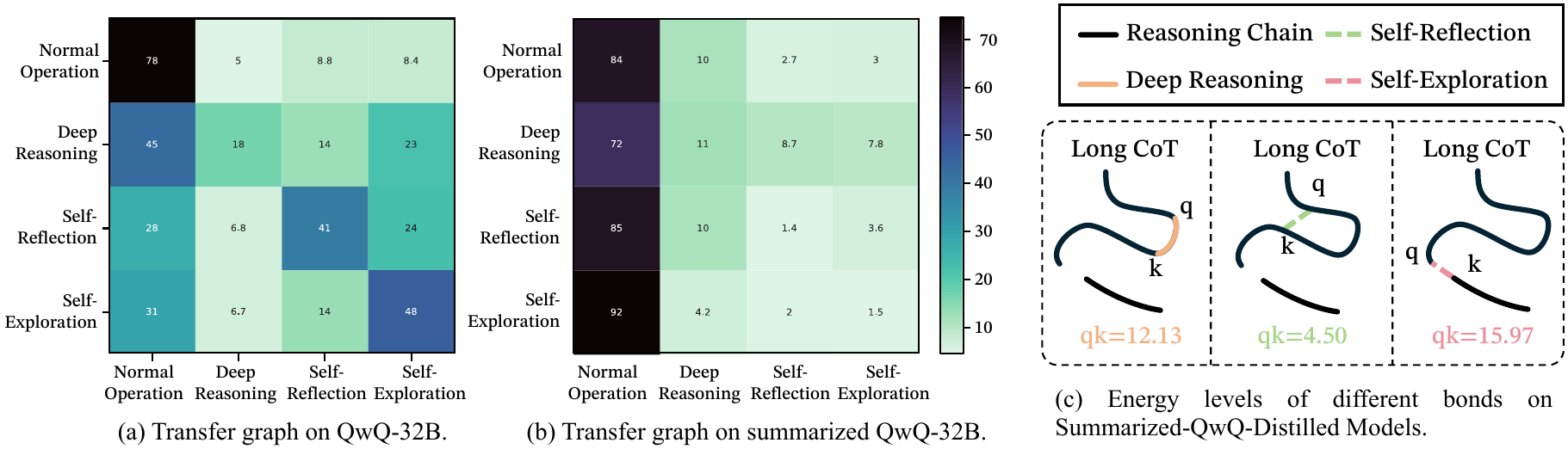}
    \caption{The reasoning behavior distribution of summarized QwQ traces.}
    \label{fig:summary}
\end{figure*}

\section{Deteriorated Molecular Structure Cannot Be Easily Restored}

\paragraph{\textbf{How Current Private LLMs Protect Their Long CoT from Distillation?}}
Exposing reasoning traces allows LLMs to imitate both answers and procedures. 
Common defenses always consider compressing intermediate steps.
We quantify this by distilling from  Gemini-2.5-Pro-Thinking and Claude-4-Sonnet. Table~\ref{tab:distill} shows that beyond $\sim$45\% token reduction versus QwQ-32B rationales, distillation causes accuracy drops, showing compression can disrupt Long CoT structure.\vspace{-8pt}

\paragraph{\textbf{Summarization break reasoning bond distributions to prohibit distillation.}}
To further validate the effectiveness of summarization, we summarized Long-CoT traces from QwQ and R1. Table~\ref{tab:summarization} shows that training on compressed trajectories yields weaker performance than training on full traces and reduces distillation effectiveness by 2\%.
In Figure~\ref{fig:summary}, summarization shifts reasoning behavior distributions and creates a gap between observable outputs and internal error-bounded transitions, limiting trace inversion and behavioral cloning. However, compression can also protect model architecture and embedded inductive priors from unauthorized imitation.

\begin{TakeawayBox}{Takeaway 6}
Summarization and reasoning compression effectively protect Long CoT structures from distillation by disrupting structural coherence, preventing unauthorized replication of internal reasoning processes.
\end{TakeawayBox}

%% file: sections/090related.tex
\vspace{-1mm}\section{Related Work}\vspace{-1mm}
Chain-of-Thought (CoT) elicits intermediate rationales, improving multi-step reasoning on math and logic tasks. Scaling short CoT to Long CoT is difficult: coherence often degrades, and cold-start gains frequently require targeted training or high-quality trajectories~\citep{guo2025deepseek,moshkov2025aimo}. 
A common approach distills stepwise solutions from strong teacher models into weaker students~\citep{muennighoff2025s1,zhao20251}. 
Outcomes hinge on reasoning quality: strong reasoning models transfer useful behaviors~\citep{ye2025limo}, whereas weaker instruction models may mimic format without robust Long CoT capability~\citep{du2025the,chandra2025shape}.

Long CoT training elicits three behaviors: deep reasoning, self-reflection, and self-exploration~\citep{o1,guo2025deepseek,chen2025towards}. Work has studied their roles; \citet{madaan2023self} used self-reflection to revise earlier steps, and \citet{shinn2023reflexion} combined reflection and exploration to improve robustness. Early studies framed CoT as sequences of these behaviors, emphasizing step-level imitation and local coherence~\citep{wei2022chain,kojima2022large,chen2024unlocking}. Later work used tree- or graph-structured reasoning to capture branching and revisitation~\citep{yao2023tree,yao2024got,hu2024treeplanner,besta2024graph}.

Although trees or graphs represent individual Long CoT traces by modeling behaviors as nodes, they do not capture the overall distribution of logical behaviors. In contrast, our approach models Long CoT as a molecular-like structure, with edges encoding stable distributions of reasoning behaviors, to test how their arrangement and interactions support effective learning.

%% file: sections/100conclusion.tex
\section{Conclusion}
This study provides a mechanistic account of Long CoT learning, conceptualized as molecular-like reasoning structures emerging through Self-Reflection, Deep-Reasoning, and Self-Exploration bonds. We propose a semantic isomer framework through behavior-transition distributions, offering insight into the stability and failure modes of Long CoT learning. Building on these, our \method leverages distribution transfer graphs to construct robust Long CoT structures, thereby improving  performance and enhancing the stability of RL.

\newpage
\section*{Limitations}
While our approach achieves strong performance on several reasoning benchmarks, it still has several limitations:
First, limited by cost and scale constraints, our analysis relies on a limited set of teacher models and student backbones, which may introduce bias in observed statistical patterns of Long CoT toward specific architectures or training recipes. Second, we focus on offline distillation and supervised fine-tuning, leaving open how well the method scales in realistic online or interactive settings with RL-like feedback. We can only approximately visualize the geometric characteristics of the inferred bond in information and semantic spaces. However, accurately delineating a universal Long CoT macromolecular structure remains an important future direction. Finally, our behavior analysis relies on an automatically labeled dataset. Even though we provide an initial robustness check,  label noise or bias will inevitably appear on estimated distributions.

%% file: sections/appendix.tex
\section{General Experimental Setting}
\paragraph{\textbf{Target Model.}} Unless otherwise specified, we start from a generic instruction-tuned LLM (e.g., Llama-3.1-8B-Instruct~\citep{dubey2024llama}). The model is standard pre-trained and supervised fine-tuned, but is not specifically optimized for long chain-of-thought reasoning.\vspace{-5pt}

\paragraph{\textbf{Data Source.}} We randomly sample 20K--35K high-quality Long CoT queries from OpenThoughts-3~\citep{guha2025openthoughts} as the primary training corpus. The dataset provides multi-step reasoning traces, averaging over 20 steps per example, and covers diverse mathematical and logical problem types.\vspace{-5pt}

\paragraph{\textbf{Evaluation Benchmarks.}} We evaluate on 6 challenging mathematical reasoning benchmarks that require multi-step logic:
\begin{itemize}[leftmargin=16pt, itemsep=0pt, topsep=0pt]
  \item \textbf{GSM8K~\citep{cobbe2021training}:} Grade-school math problems requiring multi-step calculations.
  \item \textbf{MATH-500~\citep{hendrycks2021measuring}:} High-school and early college-level problems across major math topics.
  \item \textbf{AMC 2023~\citep{amc2023}:} High-school competition problems in algebra, geometry, and combinatorics.
  \item \textbf{AIME 2024~\citep{aime2024}:} Invitational-level problems with integer final answers.
  \item \textbf{AIME 2025~\citep{aime2025}:} Invitational-level problems with integer final answers.
  \item \textbf{OlymBench~\citep{he2024olympiadbench}:} A comprehensive Olympiad-level benchmark.
\end{itemize}
Unless stated otherwise, (Overall) Accuracy is the mean accuracy across all benchmarks above.\vspace{-5pt}

\paragraph{\textbf{Inference and Metrics.}}
For evaluation, we use the same sampling temperature as in reinforcement learning, \(=0.6\), to reduce the SFT--RL mismatch. We report Avg@1 accuracy for GSM8K, MATH-500, and OlymBench. For smaller test sets (AMC 2023 and AIME 2024/2025), we report Avg@16 accuracy. We instruct the model to output the final answer in $\backslash$boxed\{$\cdot$\} format to simplify extraction. We then parse boxed answers and match them to reference solutions using standard answer-matching tools (e.g., \texttt{math-verify}). Except for specifically specified variables, all model training settings are the same.

\section{Detailed Experimental Settings for Preliminary Study}
\label{append:preliminary-data}
In this appendix, we detail the experimental setup for the preliminary study on cold-start Long CoT.
We compare 3 data construction pipelines. For all pipelines, we fine-tune the model for 1 epoch with a learning rate of $2e-5$ and a global batch size of $128$, using a max  sequence length of $16K$ or $32K$ tokens (we select the better-performing setting between these two lengths). We use Qwen2.5~\citep{yang2024qwen25} and Llama3.1~\citep{dubey2024llama} as backbones to cold-start Long CoT across multiple model sizes.\vspace{-5pt}

\begin{figure*}[t]
    \centering
    \includegraphics[width=\textwidth]{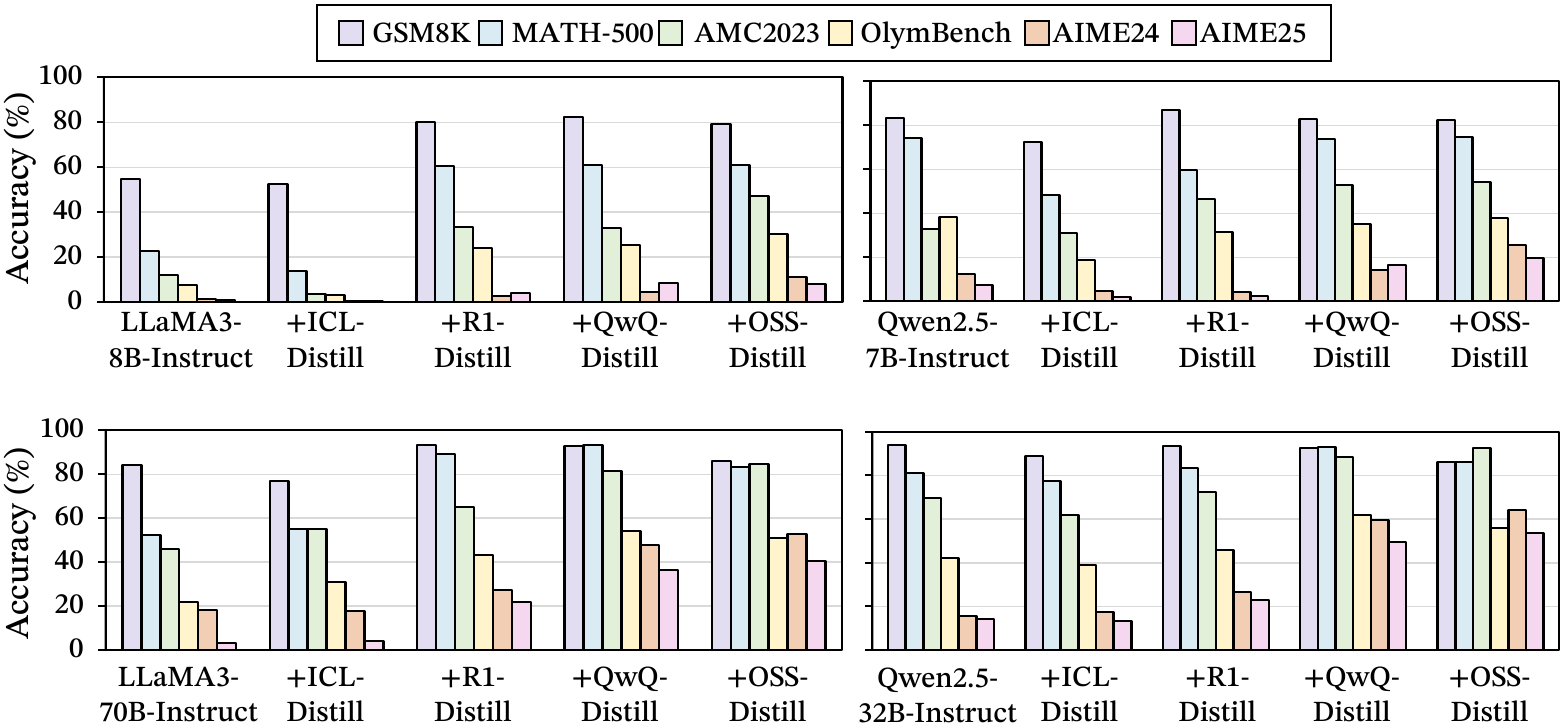}
    \caption{The full failure result of ICL from weak instruction-following LLMs and Human-annotated reasoning traces to acquire Long CoT structures, compared to successful distillation from strong reasoning LLMs.}
    \label{fig:comparison-full}
\end{figure*}
\paragraph{\textbf{Setting 1: Distillation from Strong Reasoning LLMs.}}
To construct a high-quality synthetic dataset, we use a reasoning LLM as the teacher. Specifically, we distill Long CoT reasoning traces from DeepSeek-R1-671B-0528~\citep{guo2025deepseek}, QwQ-32B~\citep{qwq2024}, OpenAI-OSS-120B~\citep{gptoss}.\vspace{-5pt}

\paragraph{\textbf{Setting 2: Distillation from Weak Instruction LLMs (ICL-Distill).}}
To simulate surface imitation via in-context learning in instruction-tuned LLMs, we use a standard instruction model (e.g., Qwen2.5-32B-Instruct) that is not optimized for deep reasoning. We randomly select 1-shot exemplar (an R1-generated Long CoT trace) in the prompt, and generate solutions for the same problems as in Setting 1.\vspace{-5pt}

\paragraph{\textbf{Setting 3: Fine-tuning on Human-Annotated Traces.}}
We utilize 50 R1-generated reasoning traces and 50 human-written step-by-step solutions for supervised fine-tuning to compare against human-annotated reasoning traces. The data are collected from \citet{du2025the}.

\section{Reasoning Bond in Long CoT}
In this section, we provide detailed experimental settings for analyzing reasoning bonds in Long CoT.

\subsection{Mathematical Definition \& Analysis of Reasoning Bonds}
\label{append:math}
\subsubsection{Reasoning Behavior Definition.}
Given an input query \(x\), a model generates an output text \(y\) containing intermediate reasoning followed by a final answer.
We segment \(y\) into \(T\) step-level units (``steps'') using standard delimiters (e.g., line breaks or bullet markers), following prior work~\citep{golovneva2023roscoe,chen2024unlocking}.
This yields a Long-CoT trajectory
\begin{equation}
\tau \coloneqq (u_1,\ldots,u_T),
\end{equation}
where each \(u_t\) is a textual step.

To analyze the geometry of \(\tau\), we map each step \(u_t\) to a vector \(h_t \in \mathbb{R}^d\).
Concretely, we encode \(u_t\) with a fixed reference encoder and obtain token hidden states from a fixed layer (e.g., the penultimate layer).
We then average token hidden states to form a step embedding \(h_t\), producing an embedding sequence \((h_1,\ldots,h_T)\).

\paragraph{Behavior space and labeled transitions.}
We model Long-CoT as a sequence of step-to-step transitions.
For each \(t \in \{1,\ldots,T-1\}\), define the directed transition (edge)
\begin{equation}
e_t \coloneqq (u_t \rightarrow u_{t+1}).
\end{equation}
Let the behavior label set be
\begin{equation}
\mathcal{B} \coloneqq \{\mathcal{N},\mathcal{D},\mathcal{R},\mathcal{E}\},
\end{equation}
where \(\mathcal{N}\) denotes Normal Operation, \(\mathcal{D}\) Deep Reasoning, \(\mathcal{R}\) Self-Reflection, and \(\mathcal{E}\) Self-Exploration.
An automated classifier (Appendix~C.1) assigns a label \(b_t \in \mathcal{B}\) to each transition \(e_t\), yielding a labeled transition sequence \((e_t,b_t)_{t=1}^{T-1}\).

We formalize a behavior-graph framework for Long CoT, in which molecular-inspired ‘bonds’ correspond to behavior-labeled edges whose distributions can be estimated, compared, and transferred across models.

\paragraph{Definition 1 (Deep Reasoning: \(\mathcal{D}\)).}
A transition \(e_t=(u_t \rightarrow u_{t+1})\) is labeled \(\mathcal{D}\) if the primary intent of \(u_{t+1}\) is to extend the reasoning chain via non-trivial inference (e.g., multi-step causal, deductive, or analogical reasoning).
Operationally, \(\mathcal{D}\) typically introduces at least one of: (i) new latent assumptions, (ii) intermediate logical variables/nodes, or (iii) derivations that go beyond direct computation or restatement.

\paragraph{Definition 2 (Self-Reflection: \(\mathcal{R}\)).}
A transition \(e_t=(u_t \rightarrow u_{t+1})\) is labeled \(\mathcal{R}\) if \(u_{t+1}\) explicitly comments on, audits, or regulates the model's own reasoning process.
This includes expressing uncertainty, revising the solution strategy, identifying or correcting mistakes, or re-examining earlier steps (i.e., tracing back to prior logical nodes).

\paragraph{Definition 3 (Self-Exploration: \(\mathcal{E}\)).}
A transition \(e_t=(u_t \rightarrow u_{t+1})\) is labeled \(\mathcal{E}\) if \(u_{t+1}\) intentionally branches into alternative hypotheses or candidate solution paths.
Operationally, \(\mathcal{E}\) raises new possibilities or sub-questions and keeps multiple paths active rather than committing to a single convergent chain.

\paragraph{Remark (Normal Operation; \(\mathcal{N}\)).}
Any transition not meeting the criteria for \(\mathcal{D}\), \(\mathcal{R}\), or \(\mathcal{E}\) is labeled \(\mathcal{N}\).
This category covers routine progression such as straightforward calculation, paraphrasing, formatting, or direct execution of an already-chosen plan.

\paragraph{Note:} It should be noted that an edge linking logical nodes does not merely connect the step $s_t$ to the next step $s_{t+1}$, but also adds edges to previously related nodes.

\subsubsection{Attention Energy Definition of Reasoning Bonds.}
We now replace the molecular analogy with a direct analysis of attention patterns.
For a given Transformer layer and head, let $q_i, k_j \in \mathbb{R}^{d_k}$ denote the query and key vectors of the $i$-th and $j$-th tokens, respectively.
The attention weight $\alpha_{ij}$ from token $i$ to token $j$ is
\begin{equation}
    s_{ij} = \frac{q_i^\top k_j}{\sqrt{d_k}},
\end{equation}
\begin{equation}
\alpha_{ij}
=
\frac{\exp \left(\frac{q_i^\top k_j}{\sqrt{d_k}}\right)}
     {\sum_{\ell}\exp \left(\frac{q_i^\top k_\ell}{\sqrt{d_k}}\right)}.
\label{eq:attn-softmax}
\end{equation}
Following the standard Gibbs–Boltzmann parametrization, we define the \emph{attention energy}
\begin{equation}
E_{ij} \triangleq -s_{ij} = -\frac{q_i^\top k_j}{\sqrt{d_k}}.
\label{eq:attn-energy-def}
\end{equation}
Then, substituting Eq.~(\ref{eq:attn-energy-def}) into Eq.~(\ref{eq:attn-softmax}) yields
\begin{equation}
\alpha_{ij}
=
\frac{\exp(-E_{ij})}{\sum_{\ell}\exp(-E_{i\ell})},
\end{equation}
which is a Boltzmann distribution over targets $j$ with inverse temperature fixed to $1$.\footnote{We do not assume any physical semantics for $E_{ij}$; it is simply a convenient reparameterization of the attention logits.}
That is, the attention weight is $E_{ij}$ Gibbs–Boltzmann distribution (temperature is taken to be 1).
In this view, lower $E_{ij}$ corresponds exactly to higher attention weight $\alpha_{ij}$, i.e., stronger dependency from $i$ to $j$.

To relate this to Long CoT behaviors, we use the edge-level labeling procedure from Appendix~C.3.
Each reasoning transition (edge) in a trajectory is classified as one of three behaviors: Deep Reasoning, Self-Reflection, or Self-Exploration (plus Normal Operation as a local baseline).
For each labeled edge $b$, we aggregate the token-level attention weights between the source and target steps into a single scalar logit, and convert it to an energy value $E_b$ using Eq.~(\ref{eq:attn-energy-def}).
This yields three random variables $E_{\mathcal{D}}, E_{\mathcal{R}}, E_{\mathcal{E}}$ corresponding to Deep, Reflection, and Exploration edges, respectively.

\subsubsection{Attention-Energy Ordering of Reasoning Bonds}
To analyze ordering among attention energies for different reasoning bonds, model queries, and keys with Rotary Positional Embedding (RoPE)~\citep{su2024roformer}.
In an autoregressive Transformer, attention weights are a softmax of pre-softmax logits; define an attention energy as the negative logit so that higher attention corresponds to lower energy (Boltzmann/Gibbs form).

\paragraph{RoPE model}
RoPE can be written as a position-dependent block-diagonal rotation on (pairs of) coordinates.
Model this head's queries/keys as
\begin{equation}
q_i = R(i)u_i,\qquad k_j = R(j)v_j,
\end{equation}
where $R(t)\in\mathbb{R}^{d_k\times d_k}$ is deterministic and orthogonal.
Then
\begin{equation}
s_{ij}
=\frac{u_i^\top R(i)^\top R(j) v_j}{\sqrt{d_k}}
=\frac{u_i^\top R(i-j)v_j}{\sqrt{d_k}},
\end{equation}
so the score depends only on the relative offset.

\paragraph{Bond random variables and expectations}
Let $d_D:=1$ (adjacent), and choose integers $1<d_R<d_E$.
Define bond-energy \emph{random variables} (random over positions/data)
\begin{equation}
\mathcal{E}_{\mathcal{D}}(i) := E_{i,i-1},\qquad
\mathcal{E}_{\mathcal{R}}(i) := E_{i,i-d_R},\qquad
\mathcal{E}_{\mathcal{E}}(i) := E_{i,i-d_E}.
\end{equation}
Define their mean energies (scalars)
\begin{equation}
\bar E_{\mathcal{D}}:=\mathbb{E}\,(\mathcal{E}_{\mathcal{D}}(i)),\quad
\bar E_{\mathcal{R}}:=\mathbb{E}\,(\mathcal{E}_{\mathcal{R}}(i)),\quad
\bar E_{\mathcal{E}}:=\mathbb{E}\,(\mathcal{E}_{\mathcal{E}}(i)).
\end{equation}
Goal: prove $\bar E_{\mathcal{D}}<\bar E_{\mathcal{R}}<\bar E_{\mathcal{E}}$ in expectation, and then with high probability for empirical averages.

\paragraph{Assumptions}
Here, we should establish weak dependence under RoPE and state two assumptions on the query/key distributions.

\noindent\textbf{A1 (Isotropic, distance-decaying cross-covariance).}
There exists a scalar function $\rho(d)\ge 0$, strictly decreasing in $d$, such that for $d=|i-j|$,
\begin{equation}
\mathbb{E}[u_i v_j^\top]=\rho(d)\,I.
\end{equation}

\noindent\textbf{A2 (Positive average alignment of rotation).}
Define
\begin{equation}
\mu(d):=\frac{1}{\sqrt{d_k}}\operatorname{tr}(R(d)).
\end{equation}
Assume $\mu(d)$ is non-increasing in $d$ and $\mu(d)\ge \underline{\mu}>0$ for $d\in\{1,d_R,d_E\}$. 

\paragraph{Theorem 1 (Expected bond-energy order under RoPE).}
Under A1--A2 and $1<d_R<d_E$,
\begin{equation}
\bar E_{\mathcal{D}} < \bar E_{\mathcal{R}} < \bar E_{\mathcal{E}}.
\end{equation}

\paragraph{Proof.}
First, compute the expected logit at distance $d$. Let $j=i-d$. Then
\begin{align}
\mathbb{E}[s_{i,i-d}]
&=\mathbb{E}\left[\frac{u_i^\top R(d)v_{i-d}}{\sqrt{d_k}}\right] \\
&=\frac{1}{\sqrt{d_k}}\operatorname{tr}\!\left(R(d)\,\mathbb{E}[v_{i-d}u_i^\top]\right).
\end{align}
By A1, $\mathbb{E}[v_{i-d}u_i^\top]=\rho(d)I$. Hence
\begin{equation}
\mathbb{E}[s_{i,i-d}] = \rho(d)\cdot \frac{1}{\sqrt{d_k}}\operatorname{tr}(R(d))=\rho(d)\mu(d).
\end{equation}
By A1, $\rho(1)>\rho(d_R)>\rho(d_E)$.
By A2, $\mu(1)\ge \mu(d_R)\ge \mu(d_E)\ge \underline{\mu}>0$.
Therefore
\begin{equation}
\mathbb{E}[s_{i,i-1}]>\mathbb{E}[s_{i,i-d_R}]>\mathbb{E}[s_{i,i-d_E}].
\end{equation}
Finally, since $E_{ij}=-s_{ij}$, taking negatives yields
\begin{equation}
\bar E_{\mathcal{D}} < \bar E_{\mathcal{R}} < \bar E_{\mathcal{E}}.
\end{equation}

\paragraph{The lower bound between two bond energies.}
Let $\widehat{s}(d)$ be the sample mean of $s_{i,i-d}$ over $N$ independent draws (or approximately independent blocks),
and define $\widehat{E}(d):=-\widehat{s}(d)$.
Let $\widehat{E}_{\mathcal{D}}:=\widehat{E}(1)$, $\widehat{E}_{\mathcal{R}}:=\widehat{E}(d_R)$, $\widehat{E}_{\mathcal{E}}:=\widehat{E}(d_E)$.

\noindent\textbf{A3 (Sub-Gaussian logits).}
For each fixed distance $d\in\{1,d_R,d_E\}$, the centered logit
$s_{i,i-d}-\mathbb{E}[s_{i,i-d}]$ is $\sigma^2$-sub-Gaussian.

Then for each $d$ and any $\epsilon>0$,
\begin{equation}
\Pr\!\left(\big|\widehat{s}(d)-\mathbb{E}[s(d)]\big|>\epsilon\right)\le
2\exp\!\left(-\frac{N\epsilon^2}{2\sigma^2}\right).
\end{equation}
By a union bound over the three distances, all three estimates concentrate simultaneously with probability at least $1-\delta$
when
\begin{equation}
\epsilon=\sigma\sqrt{\frac{2\log(6/\delta)}{N}}.
\end{equation}

If the expected logit gaps satisfy
\begin{equation}
\mathbb{E}[s(1)]-\mathbb{E}[s(d_R)] > 2\epsilon,\qquad
\mathbb{E}[s(d_R)]-\mathbb{E}[s(d_E)] > 2\epsilon,
\end{equation}
then with probability at least $1-\delta$, that satisfies
$\widehat{E}_{\mathcal{D}} < \widehat{E}_{\mathcal{R}} < \widehat{E}_{\mathcal{E}},$
since $\widehat{E}(d)=-\widehat{s}(d)$ preserves strict inequalities after flipping signs.

\paragraph{Lemma (Finite-sample ordering with high probability).} 
Suppose that the mean scores satisfy \(\Delta_{DR} = \mathbb{E}[s_D] - \mathbb{E}[s_R] > 0\) and \(\Delta_{RE} = \mathbb{E}[s_R] - \mathbb{E}[s_E] > 0\). For any \(0 < \epsilon < \min(\Delta_{DR}, \Delta_{RE})/2\), if
\begin{equation}
N \ge \frac{2\sigma^2}{\epsilon^2} \log \frac{4}{\delta},
\end{equation}
then, with probability at least $1-\delta$, the empirical estimates follow the same ordering:
\begin{equation}
\Pr(\hat{E}_D < \hat{E}_R < \hat{E}_E) \ge 1-\delta.
\end{equation}

\subsubsection{Low-energy edges dominate path aggregation.}
\paragraph{Step-level dependency graph.}
Let a model-produced chain-of-thought be partitioned into reasoning steps.
Represent these steps as nodes in a directed graph \(G=(V,B)\).
For two steps \(u,v\in V\), include a directed edge \(b=(u\to v)\in B\) if step \(v\) attends to step \(u\) with average attention weight above a fixed threshold.
(Thus, edges point from an attended-to step to the attending step.)

\paragraph{Step-level edge energy.}
Assume token-level energies \(E_{ij}\) are defined for token pairs \((i,j)\).
For an edge \(b=(u\to v)\), define the set of cross-step token pairs
\begin{equation}
\mathcal{C}(u\to v) \;=\; \{(i,j): i\in \text{tokens}(u),\; j\in \text{tokens}(v)\}.
\end{equation}
Define the step-level edge energy as the mean token-level energy across these pairs:
\begin{equation}
E_b \;\equiv\; E_{u\to v}
\;=\;
\frac{1}{|\mathcal{C}(u\to v)|}\sum_{(i,j)\in \mathcal{C}(u\to v)} E_{ij}.
\end{equation}

\paragraph{Reasoning paths and path energy.}
A reasoning path \(p\) from a source step \(s\) to a target step \(t\) is a sequence of edges
\begin{equation}
p=(b_1,\dots,b_L),\quad b_\ell=(u_\ell\to u_{\ell+1}),\quad u_1=s,\quad u_{L+1}=t.
\end{equation}
Define the path energy as the additive cost
\begin{equation}
\mathcal{E}(p) \;=\; \sum_{\ell=1}^{L} E_{b_\ell}.
\end{equation}

\paragraph{Soft-min (effective) energy over all paths.}
Let \(\mathcal{P}(s\to t)\) denote the set of all paths from \(s\) to \(t\) in \(G\).
Following the standard log-sum-exp (soft-min) construction, define the effective energy
\begin{equation}
\mathcal{E}^\star(s\to t)
\;=\;
-\log \sum_{p\in\mathcal{P}(s\to t)} \exp \bigl(-\mathcal{E}(p)\bigr).
\end{equation}
Equivalently, each path receives an (unnormalized) Gibbs weight \(\exp(-\mathcal{E}(p))\), so lower-energy paths contribute more strongly to the aggregate dependence from \(s\) to \(t\).

\paragraph{Low-energy edges define effective constraints.}
\label{prop:low-energy-constraints}
Consider two paths \(p\) and \(p'\) from \(s\) to \(t\) of equal length \(L\).
Assume they are identical except at a single position \(\ell^\ast\), where \(p\) uses edge \(b_{\ell^\ast}\) and \(p'\) uses \(b'_{\ell^\ast}\).
If \(E_{b_{\ell^\ast}} \le E_{b'_{\ell^\ast}} - \delta\) for some \(\delta>0\), then
\begin{equation}
\frac{\exp \bigl(-\mathcal{E}(p)\bigr)}{\exp \bigl(-\mathcal{E}(p')\bigr)}
=
\exp \bigl(\mathcal{E}(p')-\mathcal{E}(p)\bigr)
\;\ge\;
\exp(\delta).
\end{equation}
Thus, holding all other edges fixed, replacing a higher-energy edge by a lower-energy edge amplifies the relative influence of that path by at least a factor \(\exp(\delta)\) in the attention-induced dependency distribution.

\paragraph{Proof.}
Because \(p\) and \(p'\) differ only at \(\ell^\ast\),
\begin{equation}
\mathcal{E}(p)-\mathcal{E}(p')
=
E_{b_{\ell^\ast}}-E_{b'_{\ell^\ast}}.
\end{equation}
The assumption \(E_{b_{\ell^\ast}} \le E_{b'_{\ell^\ast}} - \delta\) implies
\begin{equation}
\mathcal{E}(p)\le \mathcal{E}(p')-\delta
\quad\Rightarrow\quad
\mathcal{E}(p')-\mathcal{E}(p)\ge \delta.
\end{equation}
Exponentiating yields
\begin{equation}
\frac{\exp \bigl(-\mathcal{E}(p)\bigr)}{\exp \bigl(-\mathcal{E}(p')\bigr)}
=
\exp \bigl(\mathcal{E}(p')-\mathcal{E}(p)\bigr)
\ge
\exp(\delta),
\end{equation}
which proves the claim.

\paragraph{Analysis.}
Proposition~\ref{prop:low-energy-constraints} shows that low-energy edges act as effective constraints: they bias the model toward reusing specific multi-step dependency patterns, because any path that swaps in a lower-energy edge gains multiplicative weight in the induced path distribution.
If edges labeled as Deep Reasoning consistently exhibit lower energies than alternative behaviors, then multi-hop dependencies that rely more heavily on Deep Reasoning edges will dominate the effective aggregation from premises to conclusions, stabilizing long chain-of-thought structure without invoking an external chemical analogy.

\subsubsection{Long CoT process is looking for a more stable reasoning structure.}

Generally speaking, in chemistry, reactions tend to synthesize more stable, lower-energy compounds. Analogously, we hypothesize that the \textbf{Long CoT learning process} seeks a stable reasoning configuration that minimizes attention energy. We now formalize this intuition under mild ergodicity and bounded-energy assumptions.

\paragraph{Step 1. Stationary behavior frequencies.}  
Assume that the reasoning behavior sequence \((s_t)_{t\ge 1}\) forms an irreducible, aperiodic, time-homogeneous Markov chain with transition matrix \(P\) and stationary distribution \(\pi\) satisfying \(P^\top \pi = \pi\). By the ergodic theorem for finite-state Markov chains, the empirical frequency of each behavior converges almost surely to its stationary probability\footnote{Note: The energy at $t$ step is related to the previous logical node (node), but has nothing to do with the logical step (edge).}:
\begin{equation}
\frac{1}{T-1}\sum_{t=1}^{T-1}\mathbf{1}[s_t=b]
\;\xrightarrow[T\to\infty]{a.s.}\;
\pi_b, \quad \forall b \in \mathcal{B}.
\end{equation}\vspace{-5pt}

\paragraph{Step 2. Decomposition of the time-averaged energy.}  
Let \(E_t\) denote the \textit{bond-level attention energy} associated with transition \(e_t=(u_t \to u_{t+1})\), as defined in Appendix~\ref{append:attention}. The trajectory-level average energy can be decomposed by behavior type as
\begin{equation}
\widehat{E}_T
=\sum_{b\in\mathcal B}
\underbrace{\Biggl(\frac{1}{T-1}\sum_{t=1}^{T-1}\mathbf{1}[s_t=b]\Biggr)}_{\text{Empirical frequency of behavior $b$}}
\cdot
\underbrace{\Biggl(\frac{\sum_{t=1}^{T-1} E_t \,\mathbf{1}[s_t=b]}{\sum_{t=1}^{T-1}\mathbf{1}[s_t=b]}\Biggr)}_{\text{Mean energy conditional on }s_t=b}.
\end{equation}
This simply states that the global average equals the weighted sum of per-behavior averages.\vspace{-5pt}

\paragraph{Step 3. Conditional convergence of energy averages.}  
Assume that \(\mathbb E[|E_t|] < \infty\) and, conditional on \(s_t=b\), the distribution of \(E_t\) depends on the past only through \(s_t\) (i.e., \(E_t \perp (s_{<t}, E_{<t}) \mid s_t\)). Under this mild conditional-independence assumption, the law of large numbers within each behavior implies that the inner mean converges to the conditional expectation
\begin{equation}
\frac{1}{N_b(T)} \sum_{t:\,s_t=b} E_t \;\to\; \mu_b
=\mathbb E[E_t\mid s_t=b].
\end{equation}
Combining this with the ergodic frequency limit from Step 1 yields
\begin{equation}
\widehat{E}_T \xrightarrow[T\to\infty]{a.s.} \sum_{b\in\mathcal B}\pi_b \mu_b, 
\end{equation}
thus establishing the ergodic low-energy equilibrium.\vspace{-5pt}

\paragraph{Step 4. Exponential routing preference from Gibbs attention.} 
At reasoning step \(t+1\), consider a fixed query token \(i\) selecting among candidate targets \(\mathcal S_t\). The model samples target \(s_j\in\mathcal S_t\) according to the standard Boltzmann (softmax) distribution:
\begin{equation}
\Pr(\mathcal S_t=s_j\mid i)
=\frac{\exp(-E_{ij})}{\sum_{s_{\ell}\in\mathcal S_t}\exp(-E_{i\ell})}. 
\end{equation}
Assume a bounded-deviation condition: for any behavior type \(b\), the corresponding attention energies lie within an interval \([\mu_b-\Delta, \mu_b+\Delta]\). Then for any two behavior classes \(b,c\) co-occurring in \(\mathcal S_t\),
\begin{equation}
\frac{\Pr(\mathcal S_t \in  b\mid i)}{\Pr(\mathcal S_t \in  c\mid i)}
\ge \exp \bigl(-(\mu_b+\Delta)+(\mu_c-\Delta)\bigr)
=\exp \bigl((\mu_c-\mu_b)-2\Delta\bigr). 
\end{equation}
Intuitively, lower-energy behaviors are exponentially favored in the routing distribution. If Deep/Reflection behaviors exhibit mean energies \(\mu_{\mathcal D},\mu_{\mathcal R}\) separated from Exploration energy \(\mu_{\mathcal E}\) by a margin \(\gamma>0\), and \(2\Delta<\gamma\), then these lower-energy transitions will dominate with exponential advantage \(\exp(\gamma-2\Delta)\).

Therefore, the model's attention mechanism inherently biases it toward stable, low-energy Long CoT reasoning structures.

\subsection{Stable Reasoning Bond Distribution in Long CoT}
\label{append:stable-bond}
\paragraph{\textbf{Reasoning Chain Generation}}
To analyze structural properties across domains, we utilize OpenThoughts-3~\citep{guha2025openthoughts}, a mixed-domain corpus comprising math, code, and scientific reasoning tasks typical of QwQ-style distillation pipelines. We prompted DeepSeek-R1-671B, OpenAI-o1, and QwQ-32B with original queries to generate Long CoT traces. Generation employed a maximum token length of \{16,384, 32,768\}, a decoding temperature $T \in [0,1]$, top-\(p = 0.95\), and standard repetition penalties, without additional sampling constraints.\vspace{-5pt}

\paragraph{\textbf{Bond Type Annotation}}
\label{append:annotation}
Following \citet{golovneva2023roscoe,chen2024unlocking}, we segmented reasoning traces into step-level units using standard delimiters (e.g., ``$\backslash$n'', ``$\backslash$n$\backslash$n'', ``. '').
We then prompted Qwen2.5-32B-Instruct to classify the logical bond for each edge between consecutive steps \((h_{t-1}, h_{t})\) into one of three types. Validation on a 200-example subset against human annotation yielded a macro-F1 score exceeding 0.85, confirming the reliability of the automated labeling applied to the full corpus.

\begin{PromptBox}{Bond Type Annotation Prompt}
You are an expert annotator. Classify the CURRENT STEP into exactly one of the following categories of reasoning/behavior:
\begin{itemize}[leftmargin=16pt, itemsep=0pt, topsep=0pt]
    \item normal operation — Straightforward, direct operations (e.g., arithmetic, factual recall, simple step-by-step logic) without introducing new logical nodes.
    \item deep reasoning — Multi-step causal, deductive, or analogical thinking that extends the reasoning chain by introducing new logical nodes or hidden assumptions.
    \item self-reflection — commenting on its own thought process (e.g., confidence, strategy, uncertainty, mistakes, or reconsideration of earlier reasoning) and tracing back to previous logical nodes.
    \item exploration — generating new possibilities, hypotheses, or questions, branching into alternative paths rather than following a single conclusion.\\
\end{itemize}

Decision rules:
\begin{itemize}[leftmargin=16pt, itemsep=0pt, topsep=0pt]
    \item[(1)] If multiple categories seem to overlap, choose the most specific match based on intent:
    \begin{itemize}[leftmargin=16pt, itemsep=0pt, topsep=0pt]
        \item If the text is about reasoning itself → self-reflection.
        \item If the text is branching or speculating → exploration.
        \item If the text is extending the reasoning chain with deeper causality or hidden steps → deep reasoning.
        \item Otherwise, if it's just direct calculation or straightforward logic → normal operation.
    \end{itemize}
    \item[(2)] Do not label based on correctness of the reasoning — only on the behavioral style of thinking.
    \item[(3)] Ignore surface complexity (e.g., long math steps may still be normal operation if they are straightforward).
    \item[(4)] If mixed, choose the dominant intent; break ties with this priority: self-reflection > exploration > deep reasoning > normal operation.\\
\end{itemize}

Output format (strict):

Return exactly one line and nothing else:

\#\#\# \textbf{Behavior:} \{normal operation $|$ deep reasoning $|$ self-reflection $|$ exploration\}
\\

PREVIOUS STEP:

\{\}\\

CURRENT STEP:

\{\}
\end{PromptBox}

\paragraph{\textbf{Transfer Distribution Analysis}}
\label{app:stable-bond-stability}
To construct the reasoning topology (Figure~\ref{fig:transfer}), we randomly subsampled \(N \in \{0.5k, 1k, 2k, 5k, 10k, 20k\}\) examples for each model-task pair. We derived reasoning transfer distributions between sequential behaviors, aggregating transition frequencies into normalized transfer graphs defined over relative step indices or semantic clusters. Stability was assessed via Pearson correlations between graphs from differing models and sample sizes. Averaged across five independent trials, correlations exceeded 0.9 for \(N > 2k\) and 0.95 when comparing sample sizes above this threshold.

\subsection{Logical Bonding-Folding Structure in Long CoT}
\label{app:folding}
To analyze the distribution and geometric characteristics of reasoning bonds, we embed all reasoning steps into a unified semantic space. This representation enables visualization of the bonding structure and supports the definition of geometric metrics to quantify folding.

\subsubsection{Geometric Embedding and Visualization of Bonding Structure}
To analyze {logical folding}, we embed complete CoT trajectories into a high-dimensional semantic space using Qwen3-8B~\citep{yang2025qwen3}. We encode each reasoning step by averaging the penultimate-layer hidden states of its constituent tokens, representing trajectories as ordered sequences \(\{h_1, h_2, \ldots, h_T\}\), where \(T\) denotes the number of steps.
We visualize folding behavior (Figure~\ref{fig:validation}) via t-SNE, employing cosine distance and 5,000 iterations with an early exaggeration of 12.0. These parameters yield stable three-dimensional projections that preserve local neighborhood structure. For clarity, Figure~\ref{fig:validation} displays only chains and logical clusters relevant to specific behaviors, omitting extraneous points to highlight bonding functions.

Moreover, we define ``logic cluster'' $c(h_t)$ by iteratively merging instances located within a distance $\alpha$ of existing clusters. To reduce variability in t-SNE embeddings, we set $\alpha$ proportional to the spread of points:
\begin{equation}
\alpha = 0.02 \times (h_{\max} - h_{\min}),
\end{equation}
where \(h_{\max}\) and \(h_{\min}\) are the maximum and minimum pairwise distances among all embedded points in the trajectory. This adaptive threshold ensures that clustering remains sensitive to the specific geometry of each reasoning trace.

\subsubsection{Geometric metrics for folding}
\label{app:folding-metrics}
We quantify the geometry of labeled edges (\((t \rightarrow t{+}1)\)) in the embedding space, where each step is represented by a vector \(h_t \in \mathbb{R}^d\). Euclidean distance is assumed unless otherwise noted.\vspace{-5pt}

\paragraph{\textbf{Self-reflection: local movement and reconnection.}}
To determine if reflective transitions revisit semantic regions, we compute two metrics. First, the {local step distance} captures instantaneous displacement:
\begin{equation}
d_t = \left\lVert h_{t+1} - h_t \right\rVert_2.
\end{equation}
Second, for reflection-labeled edges, we measure the {return distance} to the trajectory history:
\begin{equation}
r_t = \min_{s < t} \left\lVert h_{t+1} - h_s \right\rVert_2.
\end{equation}
A reflection step is defined as {reconnecting} if the prior state minimizing \(r_t\) satisfies \(r_t < \alpha\), where \(\alpha\) is the predefined cutoff. Under this criterion, \(81.72\%\) of reflections return to consolidated clusters rather than drifting to novel areas.\vspace{-5pt}

\paragraph{\textbf{Deep reasoning: path length and cluster-level proximity.}}
Deep reasoning often involves extensive computation that appears local in endpoints but traverses a complex intermediate path. We capture this via the {geometric trajectory length}. 
This metric distinguishes direct transitions from extended traversals. Additionally, we quantify semantic movement using the {cluster-level graph distance}:
\begin{equation}
  g_t = \mathrm{dist}_{\mathcal{G}} \left(c(h_t), c(h_{t+1})\right),
\end{equation}
where \(\mathrm{dist}_{\mathcal{G}}\) computes the shortest distance between clusters in a graph \(\mathcal{G}\) whose nodes are clusters and edges connect clusters with pairwise distances below \(\alpha\).
Empirically, \(72.56\%\) of deep reasoning steps satisfy \(d_t < g_t < 3\), suggesting that deep reasoning typically progresses through nearby logical neighborhoods rather than jumping between remote regions.\vspace{-5pt}

\paragraph{\textbf{Self-Exploration: novelty and sustained drift.}}
Self-Exploration entails movement into unvisited regions. We quantify this via {novelty distance}, defined identically to displacement:
\begin{equation}
  n_t = d_t = \left\lVert h_{t+1} - h_t \right\rVert_2,
\end{equation}
where high values of \(n_t\) implies divergence from previously visited states. To distinguish transient jumps from sustained excursions, we examine the trajectory length \(\ell_t\). Self-exploration steps exhibit an average length of \(5.32\), consistent with extended traversals rather than local fluctuations.

\subsection{Attention Energy Levels of Different Logical Bonds}
\label{append:attention}

To analyze attention energy across bond types, we extract attention weights from distilled Llama-3.1-8B-Instruct fine-tuned on QwQ-generated data, using full-precision inference on reasoning queries drawn from evaluation benchmarks. For each generated step, we annotate the reasoning behavior and record attention weights from all layers and heads over preceding tokens.

Unless otherwise stated, we analyze the final attention layer, averaging across all heads to enhance the signal-to-noise ratio. We define attention metrics for specific bond types as follows: for self-reflection bonds, we measure attention where queries ($q$) from the reflection step attend to keys ($k$) in the nearest prior step within the hidden space, identified by minimizing the $L_2$ embedding distance. For deep reasoning bonds, we track attention from the current step to the immediately preceding step. Finally, for self-exploration bonds, we measure attention from the previous step directed toward the exploration step.\vspace{-5pt}

For each logical bond connecting step \(i\) and step \(j\), we construct bond-level statistics from token-level attention (equivalently, token-level energies). First, we select the final token(s) of step \(j\) as the target tokens (i.e., the tokens whose outgoing attention is used to characterize how step \(j\) relies on prior content);
Then, for each selected target token in step \(j\), we extract its attention weights to the final token of step \(i\), and then aggregate across attention heads using the head-wise mean.
We convert the aggregated attention logits to energies via Eq.~\ref{eq:energy}, and treat the resulting value as the empirical energy for that bond instance.

Applying this procedure across all bonds yields empirical distributions of attention weights and effective energies for reflection, deep-reasoning, and exploration bonds.

\subsection{SFT Learning of Bond Structures}
\label{append:sft-learning}
For the SFT analysis, we consider Llama-3.1-8B-Base that is pre-trained but not instruction-tuned on Long CoT data;
Then, we consider Llama-3.1-8B-Base trained on R1-distilled data as Think-SFT model obtained by supervised fine-tuning enriched with Long CoT traces.\vspace{-5pt}

\subsubsection{Setting: How does SFT actually learn these bond structures?}
\paragraph{\textbf{Cross-coder Sparse Auto-Encoder (SAE) Architecture.}}
\label{app:sft-sae-arch}
To analyze representational geometry, following \citet{jiralerspong2025model,lindsey2025sparse}, we train a cross-coder sparse auto-encoder (SAE). For each token, the SAE input is the concatenation of hidden states from the base model and the SFT model at aligned token positions.
Specifically, the SAE comprises an encoder and a decoder. The encoder is a single linear layer that maps the concatenated hidden state to a sparse-dimensional latent space. We induce sparsity in the latent code using an $\ell_1$ penalty, calibrated to yield an average activation rate of approximately 1\%–3\% per latent unit. The decoder is a linear layer trained to reconstruct the original concatenated hidden state from the sparse latent code.

We compute token-level feature activations by applying the encoder to each token’s concatenated hidden state. We report only features whose activation probability in the base model is more than 3$\times$ that in the SFT model, or vice versa. Using these features, we manually identify those associated with Long CoT behavior. Specifically, we label {think tokens} as tokens belonging to explicit reasoning segments in the SFT corpus (as opposed to prompt context or final-answer segments), and we select the features with the strongest enrichment on think tokens. This procedure yields a compact set of discourse-control features preferentially expressed during Long CoT reasoning. Inspecting tokens with the largest contributions to these features reveals strong associations with connective markers such as Maybe'', But/so'', and ``Alternatively''.\vspace{-5pt}

\paragraph{\textbf{Keyword manipulation dataset construction.}}
\label{app:sft-sae-keyword}
We examine whether supervised fine-tuning learns the underlying reasoning structure in the data, rather than exploiting superficial lexical cues associated with particular keywords. To this end, we construct two modified versions of the QwQ distillation training corpus. In both versions, whenever a selected keyword occurs (e.g., “wait,” “maybe,” “however”), we replace it with a meaning-preserving alternative while maintaining the local syntax and the intended progression of the reasoning.
In the first modified dataset (Keyword-variant 1), each occurrence of a target keyword is randomly replaced by one of four semantically similar alternatives, for example “wait” $\rightarrow$ “hold on,” “maybe” $\rightarrow$ “perhaps,” and “however” $\rightarrow$ “on the other hand.” In the second modified dataset (Keyword-variant 2), we apply the same procedure but use a different set of replacements than those in Keyword-variant 1. This design changes surface realizations while aiming to preserve the underlying reasoning trajectory, enabling a controlled test of sensitivity to lexical form.

For reproducibility, we summarize the replacement scheme in a table that lists, for each original keyword, the set of replacements used in Keyword-variant 1 and the distinct set used in Keyword-variant 2, formatted as: original keyword / replacement plan 1 / replacement plan 2.

The keywords and corresponding replacements for the Deep Reasoning bond are provided in the boxes below:
\begin{PromptBox}{Keywords for Deep Reasoning}

otherwise / if not / or else

therefore / thus / hence

because / since / due to the fact that

so / in that case / that means

first / to start / firstly

next / then / after that

finally / in the end / at last

then / in that case / as a consequence

note / keep in mind / remember

notice that / take note / bear in mind 

important / crucial / significant

actually / in fact / really

basically / essentially / fundamentally 

think step by step / work through it step by step / go stepwise

let's reason through this / let's work through this / let's think it through

carefully / with care / meticulously

logically / coherently / consistently

rigorously / systematically / by strict logic

assumption / premise / starting assumption

constraint / restriction / limitation

it implies that / it means that / it entails that

key insight / central idea / core insight

break it down / decompose it / split it up

\end{PromptBox}

The keywords and corresponding replacements for the Self-Reflection bond are listed in the box below:

\begin{PromptBox}{Keywords for Self-Reflection}
wait, / hold on, /  let's slow down,

but / yet / though

however / nevertheless / yet

reflect /  think back / pause to consider

verify / confirm / validate

double-check / recheck / verify again

reflection / introspection / self-examination
\end{PromptBox}
\begin{PromptBox}{Keywords for Self-Reflection (Continued)}
introspect / look inward / self-examine

I might be wrong / I could be mistaken / I may be off

I could be in error / I might be misreading this / I may be overlooking a detail

I'm not sure / I'm uncertain / I'm not certain

I'm not fully confident / I can't say with certainty / I have doubts

confidence / certainty / assurance

credence / confidence level / subjective probability

revise / adjust / update

modify / refine / rework

reconsider / rethink / take another look

check my assumptions / test my assumptions / validate my premises

audit my premises / question my starting points / recheck my presuppositions

self-critique / self-review / self-audit

self-correction / self-check / critical reflection

let me check / let me verify / let me double-check

alternatively / as an alternative / as another option

instead / in lieu of that / in place of that

conversely / on the flip side / the other way around

I'm struggling with / I’m wrestling with / I’m having a hard time with

\end{PromptBox}

The keywords and corresponding replacements for the Self-Exploration bond are listed in the box below:

\begin{PromptBox}{Keywords for Self-Exploration}

maybe / perhaps / might be

now / at this point / right now

let's / let us / we can

probably / likely / presumably

seems / appears / looks like

maybe not / perhaps not / possibly note

I’ll / we’ll / it helps to

consider/ think about / look at

assume / suppose / let's say

if / provided that / in case

explore / look into / examine

probe / dig into / unpack

consider two cases / split into two cases / handle two scenarios

self-exploration / personal exploration / inner exploration

self-discovery / discovering myself / learning about myself

values / principles / priorities

I want / I’d like / I’m aiming to
\end{PromptBox}
Both datasets preserve the same underlying trajectories and labels as the original SFT corpus. We further ensure that the distributions of problems, answer types, and trajectory lengths remain unchanged.

\begin{table*}[t]
\centering
\resizebox{0.92\textwidth}{!}{
\begin{tabular}{lcccccccc}
\toprule
    \textbf{Model} & \textbf{GSM8K} & \textbf{MATH-500} & \textbf{AIME2024} & \textbf{AIME2025} & \textbf{AMC2023} & \textbf{OlympiadBench} & \textbf{AVG} \\
    \midrule
    LLaMA-3.1-8B-Base & 7.58 & 3.20 & 0.00 & 0.00 & 4.22 & 1.19 & 2.70 \\
    \texttt{\ \ + 20K R1-Distill-Data} & 63.38 & 30.60 & 0.21 & 0.42 & 14.22 & 8.30 & 19.52 \\
    \texttt{\ \ + 20K OSS-Distill-Data} & 75.89 & 54.20 & 4.38 & 6.46 & 37.34 & 23.85 & 33.69 \\
    \texttt{\ \ + 20K QwQ-Distill-Data} & 64.53 & 32.20 & 2.92 & 0.42 & 16.72 & 8.89 & 20.95 \\
    \midrule
    Llama-3.1-8B-Instruct & 75.89 & 35.20 & 4.17 & 1.04 & 23.59 & 12.00 & 25.32 \\
    \texttt{\ \ + 20K R1-Distill-Data}& 79.91 & 60.60 & 2.50 & 3.96 & 33.13 & 23.85 & 33.99 \\
    \texttt{\ \ + 20K OSS-Distill-Data} & 79.00 & 60.80 & 10.83 & 7.71 & 47.03 & 30.22 & 39.27 \\
    \texttt{\ \ + 20K QwQ-Distill-Data} & 82.41 & 60.80 & 4.38 & 8.33 & 32.97 & 25.48 & 35.73 \\
    \midrule
    Qwen-2.5-7B-Base & 40.18 & 34.20 & 5.42 & 0.83 & 26.72 & 17.33 & 20.78 \\
    \texttt{\ \ + 20K R1-Distill-Data}& 76.14 & 24.20 & 1.20 & 2.29 & 10.00 & 5.33 & 19.86 \\
    \texttt{\ \ + 20K OSS-Distill-Data}& 84.99 & 68.40 & 6.04 & 8.13 & 46.25 & 27.70 & 40.25 \\
    \texttt{\ \ + 20K QwQ-Distill-Data}& 78.39 & 46.80 & 2.71 & 1.46 & 9.84 & 5.93 & 24.19 \\
    \midrule
    Qwen-2.5-7B-Instruct & 83.24 & 74.00 & 12.50 & 7.08 & 22.66 & 38.07 & 39.59 \\
    \texttt{\ \ + 20K R1-Distill-Data} & 87.04 & 74.80 & 14.17 & 8.54 & 46.25 & 41.48 & 45.38 \\
    \texttt{\ \ + 20K OSS-Distill-Data} & 82.31 & 74.60 & 25.42 & 19.38 & 54.34 & 37.63 & 48.94 \\
    \texttt{\ \ + 20K QwQ-Distill-Data} & 85.75 & 73.80 & 13.96 & 16.25 & 52.97 & 35.11 & 46.31 \\
    \midrule
    Qwen-2.5-32B-Base & 53.68 & 33.40 & 9.17 & 2.29 & 35.63 & 15.85 & 25.00 \\
    \texttt{\ \ + 20K R1-Distill-Data} & 76.14 & 24.20 & 1.46 & 2.29 & 10.00 & 5.33 & 19.90 \\
    \texttt{\ \ + 20K OSS-Distill-Data} & 89.76 & 77.20 & 19.38 & 17.50 & 60.63 & 39.26 & 50.62 \\
    \texttt{\ \ + 20K QwQ-Distill-Data} & 91.59 & 82.40 & 19.75 & 19.62 & 72.37 & 41.33 & 54.51 \\
    \midrule
    Qwen-2.5-32B-Instruct & 93.71 & 81.00 & 15.63 & 14.17 & 69.84 & 42.22 & 52.76 \\
    \texttt{\ \ + 20K R1-Distill-Data} & 93.63 & 83.60 & 26.67 & 22.71 & 72.56 & 45.93 & 57.52 \\
    \texttt{\ \ + 20K OSS-Distill-Data} & 86.35 & 86.20 & 64.17 & 53.54 & 92.34 & 55.70 & 73.05 \\
    \texttt{\ \ + 20K QwQ-Distill-Data} & 92.65 & 93.20 & 59.38 & 49.38 & 88.44 & 61.93 & 74.16 \\
    \midrule
    LLama-3.1-70B-Base & 46.78 & 31.80 & 3.33 & 1.88 & 32.50 & 13.19 & 21.58 \\
    \texttt{\ \ + 20K R1-Distill-Data} & 73.62 & 20.40 & 1.25 & 2.08 & 8.75 & 4.44 & 18.42 \\
    \texttt{\ \ + 20K OSS-Distill-Data} & 89.39 & 76.80 & 16.88 & 17.29 & 57.97 & 36.74 & 49.18 \\
    \texttt{\ \ + 20K QwQ-Distill-Data} & 88.86 & 82.40 & 19.58 & 18.33 & 72.34 & 39.41 & 53.49 \\
    \midrule
    LLama-3.1-70B-Instruct& 84.23 & 52.20 & 17.92 & 3.13 & 45.63 & 21.63 & 37.45 \\
    \texttt{\ \ + 20K R1-Distill-Data}& 94.62 & 80.60 & 27.29 & 21.88 & 64.84 & 43.11 & 55.39 \\
    \texttt{\ \ + 20K OSS-Distill-Data}& 85.75 & 83.40 & 52.50 & 40.42 & 84.38 & 50.81 & 66.21 \\
    \texttt{\ \ + 20K QwQ-Distill-Data}& 93.33 & 89.00 & 47.50 & 36.25 & 81.41 & 54.07 & 66.93
    \\
\bottomrule
\end{tabular}
}
\caption{Full results on GSM8K, MATH-500, AIME2024, AIME2025, AMC2023, and OlympiadBench.}
\label{tab:full_distill_result}
\end{table*}

\section{Semantic Isomer Construction Details}
\label{sec:appendix_isomers}
This section details the construction and analysis of semantic isomers across four key dimensions. We first describe the distillation of well-structured semantic isomers and their simulation via in-context learning (ICL). Subsequently, we examine information flow through metacognitive oscillation analysis and conclude with the specifics of conflict learning between two stable structures.

\subsection{Distillation of Well-structured Semantic Isomers}
The setting is the same as in Appendix~\ref{append:preliminary-data}. We use 8 base and instruct LLMs as the backbone and distill from 3 advanced reasoning LLMs.

\subsection{ICL Simulation of Semantic Isomer Structures}
\label{sec:appendix_icl}

Inspired by \citet{dong2024survey}, we study whether demonstration selection in an in-context learning (ICL) setting can approximate the effective semantic isomer structure of a target teacher model (Qwen2.5-32B).\vspace{-5pt}

\paragraph{\textbf{Demonstration Construction.}}
We built a candidate pool by using QwQ-32B to generate Long-CoT solutions for the training questions in Appendix~\ref{append:preliminary-data}.\vspace{-5pt}

\paragraph{\textbf{Demonstration Selection.}}
For each target question, we select a 1-shot demonstration by comparing the reasoning-key distribution of each candidate trace with that of the target teacher (Qwen2.5-32B) trace for the same question, using Pearson correlation. We consider three strategies: (1) \textit{Random:} Sample a demonstration uniformly from the pool. (2) \textit{High-correlation (Aligned):} Choose a demonstration with high correlation (\(r \gtrsim 0.9\)) to the teacher distribution. (3) \textit{Low-correlation (Mismatched):} Choose a demonstration with low correlation (\(r < 0.8\)), corresponding to structurally incompatible reasoning paths.\vspace{-5pt}

\paragraph{\textbf{ICL-based Distillation}}
Using the selected 1-shot demonstration, we prompt Qwen2.5-32B-Instruct to generate Long-CoT solutions for the same training questions (Appendix~\ref{append:preliminary-data}). We then fine-tune a student model (e.g., Llama3.1-8B-Instruct) on these ICL-generated traces with the same hyperparameters as in Appendix~\ref{append:preliminary-data}. For each question, we keep the demonstration fixed across all samples.

\subsection{Information Flow and Metacognitive Oscillation Analysis}
\label{append:information}

\subsubsection{Setting: Information Flow Analysis and Metacognitive Oscillation Quantification}

\paragraph{\textbf{Information Flow Analysis in Phase Space (Human vs. R1)} }
We compare the reasoning dynamics of humans and the R1 model in an information phase space by quantifying information gain, entropy evolution, and the rate of convergence.
We use multi-step logical deduction and structured reasoning tasks from \citet{du2025the}. For human data, participants wrote their reasoning process step-by-step. For model data, R1 was prompted to produce step-by-step reasoning. We treat each step as a discrete point along a reasoning trajectory. For humans, steps are segmented using natural paragraph breaks or explicit logical transitions identified by the delimiter ``$\backslash n\backslash n$''. For the model, steps are delineated using explicit reasoning markers in the generated output. For visualization, we report only three representative patterns of information-flow change.

Each reasoning step \(s_t\) is mapped to a semantic probability representation \(p_t\) using a unified semantic probability encoder (Llama-3.1-8B-Instruct). We define step-wise entropy through the tuple \((I_t, \Delta I_t)\), where \(I_t\) denotes the cumulative entropy (or cumulative information measure) up to step \(t\), and \(\Delta I_t = I_t - I_{t-1}\) denotes the instantaneous change. The full reasoning chain thus forms a trajectory in a 2D phase space with coordinates (\(I_t, \Delta I_t\)).

To characterize reasoning dynamics (e.g., uniform gain versus accelerating convergence), we analyze the distribution of  \(\Delta I_t\) and the local phase-space slope, computed between consecutive points as:
\begin{equation}
  m_t = \frac{\Delta I_t - \Delta I_{t-1}}{I_t - I_{t-1}}.
\end{equation}
when $I_t \neq I_{t-1}$. The slope \(m_t\) captures the rate of change in information gain relative to cumulative information, indicating whether the reasoning process is accelerating, decelerating, or stable at each step.\vspace{-5pt}

\paragraph{\textbf{Analysis of Metacognitive Oscillation}}
We formalize and quantify metacognitive oscillation in LLMs—alternation between high-entropy exploration and low-entropy validation—by relating these dynamic states to the distributions of reasoning bonds.
Using the phase-space dynamics, we classify each reasoning step into one of two states. (1) \textbf{High-entropy Exploration:} characterized by a steep phase-space slope (\(m_t >0.6\)) together with a substantial entropy increase (\(\Delta\text{entropy} > 0.05\)).
(2) \textbf{Low-entropy Validation:} characterized by a near-zero or negative slope (\(m_t \approx 0\)) and and minimal entropy change (\(|\Delta\text{entropy}| < 0.05\)).
We then measure the frequency and periodicity of transitions between these states, and analyze which reasoning keys (bonds) are most prevalent within each state.

\subsection{Details about Conflict Learning Between Two Stable Structures}
\label{sec:appendix_conflict}

\paragraph{\textbf{Setup about performance analysis on different training data mixture strategies.}}
To test whether simultaneously training a model on two highly correlated (\(r \approx 0.9\)) yet structurally distinct reasoning frameworks (from R1 and OSS) leads to ``structural chaos.''
Specifically, we use OpenAI-OSS-120B~\citep{gptoss} as the OSS model and DeepSeek-R1-671B-0528~\citep{guo2025deepseek} as the R1 model to generate two sets of Long CoT traces on the same 20K training questions from OpenThoughts-3~\citep{guha2025openthoughts}.
Based on these generated Long CoT traces, we created three training configurations: (1) OSS-Distill-Data, an R1-only set with 20K samples generated from DeepSeek-R1, (2) R1-Distill-Data, an OSS-only set with 20K samples generated from OpenAI-OSS, and (3) R1-then-OSS, a sequential set with first 10K from DeepSeek-R1 and then 10K from OpenAI-OSS. (4) OSS-then-R1, a sequential set with first 10K from OpenAI-OSS and then 10K from DeepSeek-R1.
(5) R1-mix-OSS, a randomly mixed set with 10K from OpenAI-OSS and 10K from DeepSeek-R1.\vspace{-5pt}

\paragraph{\textbf{Pearson correlation coefficient
between transfer distribution.}}
To quantify the similarity between two reasoning structures (e.g., R1 and OSS), we compute the Pearson correlation coefficient \(r\) between their transfer distributions. Given two transfer matrices \(P\) and \(Q\) representing the normalized frequencies of reasoning bonds between steps, we flatten these matrices into vectors \(\mathbf{p}\) and \(\mathbf{q}\). The Pearson correlation coefficient is then calculated as:
\begin{equation}
r = \frac{\sum_{i=1}^{n} (p_i - \bar{p})(q_i - \bar{q})}{\sqrt{\sum_{i=1}^{n} (p_i - \bar{p})^2} \sqrt{\sum_{i=1}^{n} (q_i - \bar{q})^2}},
\end{equation}
where \(\bar{p}\) and \(\bar{q}\) are the mean values of vectors \(\mathbf{p}\) and \(\mathbf{q}\), respectively, and \(n\) is the number of elements in each vector. A high positive correlation (close to 1) indicates similar reasoning structures, while a low or negative correlation indicates dissimilar structures.

\section{Details about Synthetic Long CoT with \method}
\label{app:settings}

This section describes the experimental settings used for synthesis, analysis, and reinforcement learning with \method.
\subsection{Supervised-Finetuning with \method}
To match the statistical properties of the synthetic chain-of-thought (CoT) to the behavioral patterns of stronger teacher models (Section~\ref{append:stable-bond}), we first estimate a reasoning-state transition distribution \(p(s_{t+1}|s_t)\). Specifically, we analyze 20k distilled CoT rationales generated by teacher models (e.g., QwQ-32B and OpenAI-OSS-120B). Each reasoning step \(s_t\) is annotated by LLMs, and we use these annotations to compute an empirical transition matrix \(\hat{P}\) (Figure~\ref{fig:transfer}). During synthesis, we initialize the process in an exploration state and then sample transitions between reasoning states according to. The prompts used for each reasoning state are listed below.

The prompt for the self-reflection state is as follows:

\begin{PromptBox}{Prompt for Self-Reflection}
Assume that you are a helpful assistant. You will receive a question and a previously reasoned rationale. If you can directly get the answer, please output the concise answer with $\backslash$boxed\{\}. Otherwise, please reflect on the response and provide a self-reflection.

Here are some reasoning behavior definitions:
\begin{itemize}[leftmargin=16pt, itemsep=0pt, topsep=0pt]
    \item normal operation — Straightforward, direct operations (e.g., arithmetic, factual recall, simple step-by-step logic) without introducing new logical nodes.
    \item deep reasoning — Multi-step causal, deductive, or analogical thinking that extends the reasoning chain by introducing new logical nodes or hidden assumptions.
    \item self-reflection — commenting on its own thought process (e.g., confidence, strategy, uncertainty, mistakes, or reconsideration of earlier reasoning) and tracing back to previous logical nodes.
    \item exploration — generating new possibilities, hypotheses, or questions, branching into alternative paths rather than following a single conclusion.
\end{itemize}

You should conduct self-reflection behavior now.\\

Please reflect on the response and provide a self-reflection.
\end{PromptBox}

The prompt for the self-exploration state is as follows:
\begin{PromptBox}{Prompt for Self-Exploration}
Assume that you are a helpful assistant. You will receive a question and a previously reasoned rationale. If you can directly get the answer, please output the concise answer with $\backslash$boxed\{\}. Otherwise, please explore a novel reasoning path in the response.\\

Here are some reasoning behavior definitions:
\end{PromptBox}
\begin{PromptBox}{Prompt for Self-Exploration (Continued)}

\begin{itemize}[leftmargin=16pt, itemsep=0pt, topsep=0pt]
    \item normal operation — Straightforward, direct operations (e.g., arithmetic, factual recall, simple step-by-step logic) without introducing new logical nodes.
    \item deep reasoning — Multi-step causal, deductive, or analogical thinking that extends the reasoning chain by introducing new logical nodes or hidden assumptions.

    \item self-reflection — commenting on its own thought process (e.g., confidence, strategy, uncertainty, mistakes, or reconsideration of earlier reasoning) and tracing back to previous logical nodes.
    \item exploration — generating new possibilities, hypotheses, or questions, branching into alternative paths rather than following a single conclusion.
\end{itemize}

You should conduct exploration behavior now.\\

  Please explore a novel reasoning path in the response.
\end{PromptBox}

The prompt for the normal-operation state is as follows:
\begin{PromptBox}{Prompt for Normal Operation}
  Assume that you are a helpful assistant. You will receive a question and a previously reasoned rationale. If you can directly get the answer, please output the concise answer with $\backslash$boxed\{\}. Otherwise, please conduct normal operation on the response.\\

Here are some reasoning behavior definitions:
\begin{itemize}[leftmargin=16pt, itemsep=0pt, topsep=0pt]
    \item normal operation — Straightforward, direct operations (e.g., arithmetic, factual recall, simple step-by-step logic) without introducing new logical nodes.
    \item deep reasoning — Multi-step causal, deductive, or analogical thinking that extends the reasoning chain by introducing new logical nodes or hidden assumptions.
    \item self-reflection — commenting on its own thought process (e.g., confidence, strategy, uncertainty, mistakes, or reconsideration of earlier reasoning) and tracing back to previous logical nodes.
    \item exploration — generating new possibilities, hypotheses, or questions, branching into alternative paths rather than following a single conclusion.
\end{itemize}

You should conduct normal operation behavior now.\\

Please conduct normal operation on the response.
\end{PromptBox}

The prompt for the deep-reasoning state is as follows:
\begin{PromptBox}{Prompt for Deep Reasoning}
  Assume that you are a helpful assistant. You will receive a question and a previously reasoned rationale. If you can directly get the answer, please output the concise answer with $\backslash$boxed\{\}. Otherwise, please further deepen the reasoning on the response.\\

Here are some reasoning behavior definitions:
\begin{itemize}[leftmargin=16pt, itemsep=0pt, topsep=0pt]
    \item normal operation — Straightforward, direct operations (e.g., arithmetic, factual recall, simple step-by-step logic) without introducing new logical nodes.
    \item deep reasoning — Multi-step causal, deductive, or analogical thinking that extends the reasoning chain by introducing new logical nodes or hidden assumptions.
    \item self-reflection — commenting on its own thought process (e.g., confidence, strategy, uncertainty, mistakes, or reconsideration of earlier reasoning) and tracing back to previous logical nodes.
    \item exploration — generating new possibilities, hypotheses, or questions, branching into alternative paths rather than following a single conclusion.
\end{itemize}

You should conduct deep reasoning behavior now.\\

Please further deepen the reasoning on the response.
\end{PromptBox}

\begin{table*}[t]
\centering
\resizebox{0.92\textwidth}{!}{
\begin{tabular}{lcccccccc}
\toprule
    \textbf{Model} & \textbf{GSM8K} & \textbf{MATH-500} & \textbf{AIME2024} & \textbf{AIME2025} & \textbf{AMC2023} & \textbf{OlympiadBench} & \textbf{AVG} \\
    \midrule
    LLaMA-3.1-8B-Base & 7.58 & 3.20 & 0.00 & 0.00 & 4.22 & 1.19 & 2.70 \\
    \texttt{\ \ + 20K Qwen-Distill-Data} & 62.47 & 29.40 & 0.00 & 0.00 & 12.81 & 6.81 & 18.58 \\
    \texttt{\ \ + 20K OSS-Distill-Data} & 75.89 & 54.20 & 4.38 & 6.46 & 37.34 & 23.85 & 33.69 \\
    \texttt{\ \ + 20K QwQ-Distill-Data} & 64.53 & 32.20 & 2.92 & 0.42 & 16.72 & 8.89 & 20.95 \\
    \texttt{\ \ + 20K OSS-\method{}} & 67.85  & 35.20  & 1.83  & 0.83  & 20.53  & 11.11  & 22.89 \\
    \texttt{\ \ + 20K QwQ-\method{}} & 66.41  & 35.00  & 2.08  & 0.63  & 20.16  & 10.37  & 22.44 \\
    \midrule
    Llama-3.1-8B-Instruct & 75.89 & 35.20 & 4.17 & 1.04 & 23.59 & 12.00 & 25.32 \\
    \texttt{\ \ + 20K Qwen-Distill-Data} & 76.50 & 39.80 & 4.38 & 1.04 & 25.63 & 19.70 & 27.84 \\
    \texttt{\ \ + 20K OSS-Distill-Data} & 79.00 & 60.80 & 10.83 & 7.71 & 47.03 & 30.22 & 39.27 \\
    \texttt{\ \ + 20K QwQ-Distill-Data} & 82.41 & 60.80 & 4.38 & 8.33 & 32.97 & 25.48 & 35.73 \\
    \texttt{\ \ + 20K OSS-\method{}} & 83.24 & 51.80 & 4.79 & 1.04 & 32.50 & 21.04 & 32.40  \\
    \texttt{\ \ + 20K QwQ-\method{}} & 84.31 & 50.20 & 5.21 & 1.67 & 32.34 & 20.00 & 32.29  \\
    \midrule
    Qwen-2.5-7B-Base & 40.18 & 34.20 & 5.42 & 0.83 & 26.72 & 17.33 & 20.78 \\
    \texttt{\ \ + 20K Qwen-Distill-Data} & 68.69 & 39.80 & 4.38 & 1.04 & 25.63 & 19.70 & 26.54 \\
    \texttt{\ \ + 20K OSS-Distill-Data} & 84.99 & 68.40 & 6.04 & 8.13 & 46.25 & 27.70 & 40.25 \\
    \texttt{\ \ + 20K QwQ-Distill-Data} & 78.39 & 46.80 & 2.71 & 1.46 & 9.84 & 5.93 & 24.19 \\
    \texttt{\ \ + 20K QwQ-\method{}} & 81.20 & 62.20 & 6.25 & 3.54 & 41.88 & 30.52 & 37.60 \\
    \texttt{\ \ + 20K OSS-\method{}} & 83.17 & 63.80 & 5.83 & 1.67 & 41.56 & 29.33 & 37.56 \\
    \midrule
    Qwen-2.5-7B-Instruct & 83.24 & 74.00 & 12.50 & 7.08 & 22.66 & 38.07 & 39.59 \\
    \texttt{\ \ + 20K Qwen-Distill-Data} & 84.31 & 63.40 & 6.46 & 3.13 & 31.72 & 29.78 & 36.46 \\
    \texttt{\ \ + 20K OSS-Distill-Data} & 82.31 & 74.60 & 25.42 & 19.38 & 54.34 & 37.63 & 48.94 \\
    \texttt{\ \ + 20K QwQ-Distill-Data} & 85.75 & 73.80 & 13.96 & 16.25 & 52.97 & 35.11 & 46.31 \\
    \texttt{\ \ + 20K QwQ-\method{}} & 89.61 & 76.00 & 7.29 & 3.96 & 51.88 & 36.74 & 44.25 \\
    \texttt{\ \ + 20K OSS-\method{}} & 88.02 & 77.80 & 8.13 & 5.00 & 52.81 & 37.48 & 44.87 \\
\bottomrule
\end{tabular}
}
\caption{Comparison of strong reasoning LLM distillation versus \method{} using a weak instructed LLM across six benchmarks.}
\label{tab:full_method_result}
\end{table*}

\begin{table*}[t]
\centering
\resizebox{\textwidth}{!}{
\begin{tabular}{lcccccccc}
\toprule
    \textbf{Model} & \textbf{GSM8K} & \textbf{MATH-500} & \textbf{AIME2024} & \textbf{AIME2025} & \textbf{AMC2023} & \textbf{OlympiadBench} & \textbf{AVG} \\
    \midrule
    
Llama-3.1-8B-Instruct	& 75.89 & 35.20 & 4.17 & 1.04 & 23.59 & 12.00 & 25.32 \\
\midrule
\texttt{\ \  + \method-by-Llama-3.1-8B-Instruct}& 36.01 & 23.60 & 2.71 & 0.21 & 13.75 & 7.41 & 13.95 \\
\texttt{\ \  + \method-by-Llama-3.1-70B-Instruct} & 82.71 & 53.60 & 3.33 & 0.63 & 31.72 & 19.85 & 31.97 \\
\texttt{\ \  + \method-by-Qwen-7B} &	83.47 & 51.60 & 3.33 & 0.63 & 32.03 & 19.85 & 31.82 \\
\texttt{\ \  + \method-by-Qwen-32B} &	82.41 & 60.80 & 4.38 & 8.33 & 32.97 & 25.48 & 35.73 \\
\bottomrule
\end{tabular}
}
\caption{The effectiveness analysis of \method{} methods based on different instruct models.}
\label{tab:different_llm_method}
\end{table*}

\begin{table*}[t]
\centering
\resizebox{0.96\textwidth}{!}{
\begin{tabular}{lcccccccc}
\toprule
    \textbf{Model} & \textbf{GSM8K} & \textbf{MATH-500} & \textbf{AIME2024} & \textbf{AIME2025} & \textbf{AMC2023} & \textbf{OlympiadBench} & \textbf{AVG} \\
    \midrule
    Llama-3.1-8B-Instruct & 75.89 & 35.20 & 4.17 & 1.04 & 23.59 & 12.00 & 25.32 \\
    \midrule
    \texttt{\ \ + 20K QwQ-Distill-Data} & 82.41 & 60.80 & 4.38 & 8.33 & 32.97 & 25.48 & 35.73  \\
    \texttt{\ \ + 20K QwQ-Distill-Data + RL} & 91.51 & 69.80 & 8.96 & 4.17 & 37.97 & 25.93 & 39.72  \\
    \midrule
    \texttt{\ \ + 20K QwQ-\method{}} & 84.31 & 50.20 & 5.21 & 1.67 & 32.34 & 20.00 & 32.29 \\
    \texttt{\ \ + 20K QwQ-\method{} + RL} & 88.78 & 70.80 & 7.50 & 3.33 & 39.22 & 21.04 & 38.44 \\
    \texttt{\ \ + 35K QwQ-\method{} + RL} & 90.30 & 68.40 & 10.00 & 4.38 & 39.84 & 24.15 & 39.51 \\
\bottomrule
\end{tabular}
}
\caption{The reinforcement learning performance based on 3 initialization methods across 6 benchmarks.}
\label{tab:rl_results}
\end{table*}

Furthermore, to investigate the impact of distinct synthetic backbones, we employed \method to distill data from various base models (Figure \ref{tab:different_llm_method}). Models with limited self-reflection or exploration capabilities, such as Llama-3.1-8B-Instruct, struggled to synthesize effective reasoning data. Conversely, combining the Qwen series with Llama-70B yielded robust results across most domains, with the notable exception of tasks requiring deep reasoning. For such tasks (e.g., AIME), performance depended heavily on the model's intrinsic reasoning depth, following the hierarchy: Qwen-32B outperforms both Qwen-7B and Llama-70B.

\subsection{Reinforcement Learning with \method Initialization}

We adopt DAPO~\citep{yu2025dapo} as the reinforcement-learning (RL) framework to fine-tune Llama-3.1-8B-Instruct models initialized from different checkpoints. To isolate the effect of initialization, we hold the RL tasks, reward functions, and all other training hyperparameters constant across experiments. Each model is trained for 1000 steps with a learning rate of \(1 \times 10^{-5}\), a batch size of 16, a sampling size of 16, and a maximum sequence length of 16384 tokens. We set the clipping parameters to clip-low $=0.2$ and clip-high $=0.68$. We use MATH~\citep{hendrycks2021measuring} and AIME 1989--2023~\citep{aime_1983_2024} as training data.

We evaluate two RL initialization strategies:
(1) \textbf{QwQ-Distill Data + RL:} RL fine-tuning starting from Llama-3.1-8B-Instruct further tuned on QwQ-generated long chain-of-thought data.
(2) \textbf{QwQ-Graph-Syn + RL:} RL fine-tuning starting from our \method-SFT model.
All runs use the same RL tasks, reward functions, and hyperparameters; therefore, performance differences can be attributed to the initialization checkpoint (i.e., the supervised pretraining/fine-tuning history).

\section{Details of Bond Shaping Function Analysis}
\label{app:bond_shaping_setting}

This section provides a detailed description of the experimental settings for the geometric reshaping analyses presented in Fig.~\ref{fig:function}, corresponding to the three "bond" types in the Long CoT structure: Deep Reasoning (covalent bond), Self-Exploration (exploration bond), and Self-Reflection (stabilizing bond).

To analyze the function of different "bonds," we extracted step-level hidden representations from the penultimate Transformer layer of both Llama-3.1-8B-Instruct and our trained \method model. For each reasoning step \(t\), we computed its vector representation \(\mathbf{h}_t\) via average pooling of its token embeddings. We then used t-SNE to visualize the semantic space and compare the representations \(\mathbf{h}^{\text{orig}}_t\) and \(\mathbf{h}^{\text{method}}_t\) for the same reasoning step across the two models.

\subsection{Deep Reasoning: Densing the Primary Structure.}
To quantify how Deep Reasoning densifies the ``primary structure,'' we use the same development-set split as in the main text. We compare two training modes: (1) \textit{baseline}, which uses standard short CoT prompting by standard Llama-3.1-8B-Instruct; and (2) \textit{deep-reasoning-trained}, which uses the Long CoT paradigm with the Deep Reasoning transfer probability set to 50\%, while the remaining probability mass follows the other behaviors in our default setting.
For each input sample, we generate 10K training instances and, under each mode, sample 100 reasoning trajectories from OlymBench. From these trajectories, we extract the step-level baseline representations \(\{{h}_1, \dots, {h}_{T}\}\) and deep-reasoning-trained representations \(\{\tilde{h}_1, \dots, \tilde{h}_{T}\}\).

To characterize the compactness of the "primary To characterize the compactness of the ``primary structure'' in semantic space, we compute the Minimum Enclosing Ball (MEB) over all step representations from the sampled trajectories:
\begin{equation}
  S = \{\tilde{h}_1, \dots, \tilde{h}_{T_{N}}\},
\end{equation}
where \(T_N\) denotes the total number of extracted steps aggregated across \(N\) trajectories for the given sample set. We compute the MEB in a shared three-dimensional embedding space obtained by t-SNE, which we use consistently across modes for visualization and for a comparable notion of geometric spread.
The procedure is as follows. First, we reduce each step representation from the original high-dimensional embedding space to \(d=3\) using t-SNE. Second, we apply an approximate MEB solver based on Welzl's algorithm to obtain the center \(c \in \mathbb{R}^{3}\) and radius \(r\). Third, we define the corresponding volume in \(d=3\) Euclidean space as:
\begin{equation}
V = C_d \cdot r^{d},
\end{equation}
where \(C_d\) is the unit-ball volume constant (a fixed scaling factor that cancels in relative comparisons).
We then compute the arithmetic mean of \(V\) over all development-set samples, denoted as \(V_{\text{base}}\) (Baseline) and \(V_{\text{deep}}\) (Deep Reasoning). The volume reduction reported in Fig.~\ref{fig:function}(a) is:
\begin{equation}
\Delta_{\text{Deep}} = \frac{V_{\text{base}} - V_{\text{deep}}}{V_{\text{base}}} \times 100\%.
\end{equation}
A positive \(\Delta_{\text{Deep}}\) indicates that Deep Reasoning contracts the semantic volume of the core logical backbone while preserving comparable semantic coverage of the task.

\subsection{Self-Exploration: Expanding the Logical Space.}
We use the same development-set split and the same two training modes as above: (1) \textit{baseline} (short CoT or direct answering, without Long CoT) and (2) \textit{self-exploration-trained} (Long CoT with self-exploration transfer probability >50\%, with other behaviors unchanged). For each input sample, we generate 10K training instances and sample 100 reasoning trajectories from OlymBench under each mode, then extract the corresponding step representations.
Let \(V_{\text{exp}}\) denote the mean MEB volume computed from the step representations used for the \textit{self-exploration-trained} model. We quantify the relative volume change as:
\begin{equation}
\Delta_{\text{Exp}} = \frac{V_{\text{exp}} - V_{\text{base}}}{V_{\text{base}}} \times 100\%.
\end{equation}
Then, a positive \(\Delta_{\text{Exp}}\) corresponds to an expansion of the explored logical space relative to the baseline.

\subsection{Self-Reflection: Densifying and Stabilizing the Logical Results.}
We follow the same data construction and representation extraction protocol as above. To measure the contraction induced by Self-Reflection, we compute MEB volumes before and after the reflection step, denoted by \(V_{\text{pre}}\) and \(V_{\text{post}}\), respectively. We then report the relative contraction as
\begin{equation}
\Delta_{\text{Reflect}} = \frac{V_{\text{pre}} - V_{\text{post}}}{V_{\text{pre}}} \times 100\%,
\end{equation}
where a larger value indicates a stronger ``folding'' effect in semantic space after reflection.

To further examine the ``suppression of inconsistent branches,'' we conduct an auxiliary clustering analysis (not shown in the main text). Specifically, we cluster the pre- and post-reflection point sets, \(S_{\text{pre}}\) and \(S_{\text{post}}\), using an \(\alpha\)-threshold grouping method. We observe that, after reflection, intra-cluster distances decrease substantially, while inter-cluster distances remain stable or increase slightly. This pattern suggests that the dominant semantic cluster becomes more compact, whereas inconsistent branches are pruned or weakened. Consistent with these findings, the overall semantic volume decreases from 35.2 to 31.2, indicating that Self-Reflection guides the structure toward a more stable solution manifold.

\begin{figure}[t]
    \centering
    \includegraphics[width=0.96\textwidth]{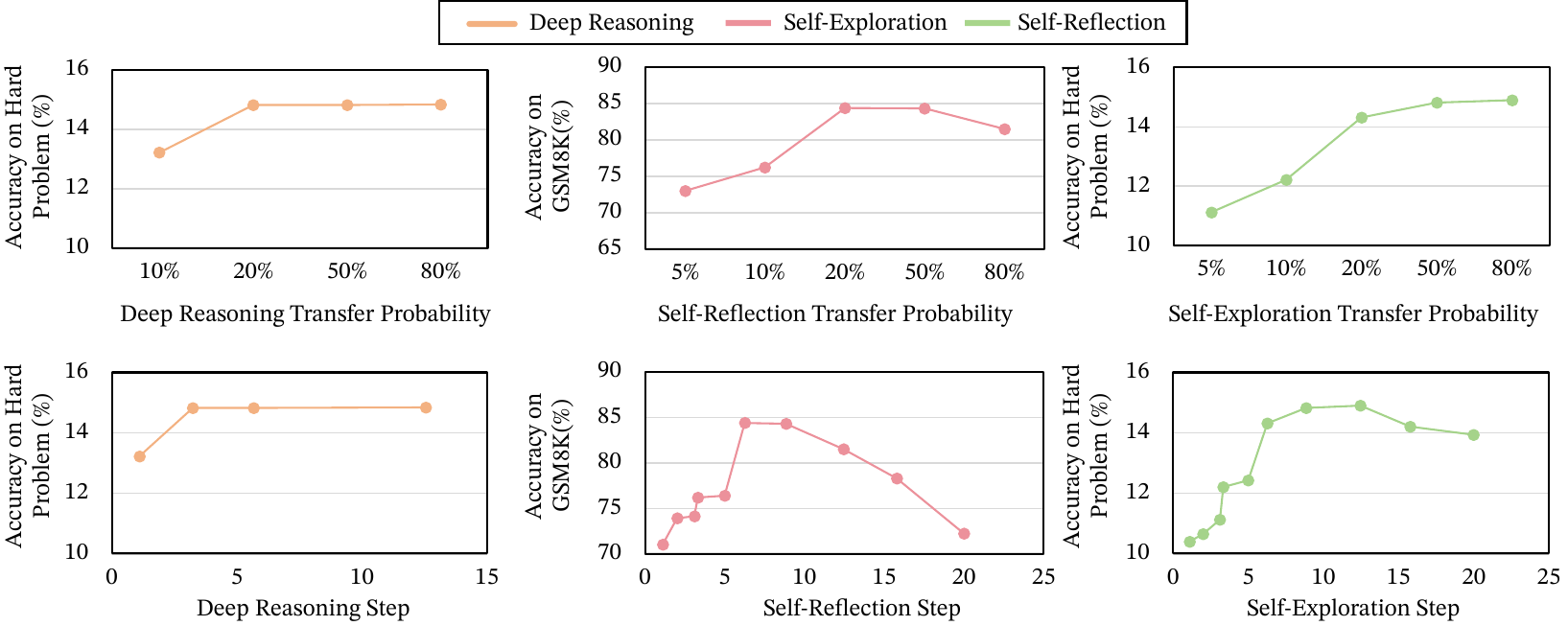}
    \caption{Comparative analysis of performance under different reasoning bond ratios.}
    \label{fig:capability}
\end{figure}
\section{Analysis of the impact of length and diversity in reasoning behaviors}

To evaluate whether comparable performance gains can be achieved merely by regulating the length and frequency of Self-Reflection steps, we conducted a detailed analysis illustrated in Fig.~\ref{fig:function}, covering the three distinct "bond" types that constitute the Long CoT framework.

\subsection{Impact of bond-enhanced behaviors on performance}

We further examined how different reasoning behaviors influence overall model performance (Fig.~\ref{fig:capability}). For complex queries, improvements are primarily driven by reasoning depth rather than the sheer number of steps, once a minimum threshold is achieved. A higher proportion of reflective behaviors tends to enhance performance on challenging tasks but can introduce “overthinking,” which reduces accuracy on simpler ones. Similarly, excessive exploratory behavior on complex tasks often impedes convergence—indeed, 62.7\% of AIME cases exhibited extended reasoning without reaching a clear conclusion. In contrast, simple queries generally benefit from more direct exploration, leading to greater accuracy. Across all models, performance consistently peaks at a balanced distribution of behaviors, suggesting the existence of a stable and task-invariant optimal configuration.

\subsection{Effect of bond length on reasoning quality}

We also investigated how varying bond lengths affects performance while maintaining constant behavioral ratios. As shown in Fig.~\ref{fig:capability}, increasing both the number of steps and the length of individual reasoning bonds generally strengthens performance, particularly in deep reasoning scenarios. However, overly long exploration bonds can induce semantic drift and reduce the precision of final outputs. The optimal bond length appears to depend on task difficulty: shorter bonds tend to perform better on simple problems, whereas longer ones are more advantageous for tackling complex challenges.

\section{Details of LLM Structure Reconstruction}
\label{app:exp_settings}

\subsection{Reasoning compression or summarization data collection.}
For the analysis about broken structure reconstruction, we treat Gemini-2.5-Pro-Thinking and Claude-4-Sonnet as black-box teacher models that can emit long chain-of-thought (CoT) traces. 
We query both APIs with identical user prompts and task instructions, and enable their ``thinking'' content, which are much shorter than other reasoning LLMs with Long CoT.\vspace{-5pt}

\subsection{Summarize Reasoning Process Analysis.}
To emulate the protection strategies used in private LLMs, we summarize the full Long CoT traces from QwQ-32B and OpenAI-OSS-120B into concise summaries by Qwen2.5-32B. This procedure removes the long-range reasoning structure of the original CoT, thereby reducing the amount of recoverable step-by-step rationale. The summarization prompt is as follows:

\begin{PromptBox}{Prompt for Summarization}
You are an expert summarizer. Below is a Long Chain-of-Thought reasoning trace generated by an AI model to solve a complex problem. Your task is to compress this reasoning process into a concise summary.\\

Input Long Chain-of-Thought Trace:

[Insert Full Trace Here]

Summary:
\end{PromptBox}
We trained the model on the summarized 20K samples. The corresponding results are reported in Table~\ref{tab:summarization}. All other analytical experiments followed the same configuration as in the preceding sections, except for the model and the reported outputs.